\documentclass[10pt,twocolumn,letterpaper]{article}

\usepackage[pagenumbers]{cvpr}

\usepackage[accsupp]{axessibility}

\definecolor{cvprblue}{rgb}{0.21,0.49,0.74}
\usepackage[breaklinks,colorlinks,allcolors=cvprblue]{hyperref}

\usepackage{threeparttable}
\usepackage{stfloats}
\usepackage{float}
\usepackage{array}
\usepackage{dsfont}
\usepackage{amsthm}
\usepackage{bm}
\usepackage{bbm}
\usepackage[capitalize,noabbrev,nameinlink]{cleveref}

\definecolor{pakistangreen}{rgb}{0.0, 0.4, 0.0}
\definecolor{bostonuniversityred}{rgb}{0.8, 0.0, 0.0}

\makeatletter

\newlength{\supptextfloatsep}
\newlength{\suppfloatsep}
\newlength{\suppintextsep}
\newlength{\suppabovecaptionskip}
\newlength{\suppbelowcaptionskip}
\newcommand{\tightpara}[1]{\par\vspace{0.35em}\noindent\textbf{#1.}\hspace{0.35em}}

\theoremstyle{plain}

\theoremstyle{definition}

\theoremstyle{remark}

\crefname{definition}{Definition}{Definitions}
\crefname{theorem}{Theorem}{Theorems}
\crefname{lemma}{Lemma}{Lemmas}
\crefname{corollary}{Corollary}{Corollaries}
\crefname{remark}{Remark}{Remarks}
\crefname{problem}{Problem}{Problems}
\crefname{fact}{Fact}{Facts}
\crefname{proposition}{Proposition}{Propositions}
\crefname{example}{Example}{Examples}
\crefname{simulation}{Simulation}{Simulations}
\crefname{listthm}{Theorem}{Theorems}
\crefalias{thmlisti}{listthm}

\crefname{section}{Section}{Sections}
\crefname{figure}{Figure}{Figures}
\crefname{table}{Table}{Tables}
\crefname{algorithm}{Algorithm}{Algorithms}
\crefname{appendix}{Appendix}{Appendices}

\DeclareFontFamily{U}{mathx}{}
\DeclareFontShape{U}{mathx}{m}{n}{<-> mathx10}{}
\DeclareSymbolFont{mathx}{U}{mathx}{m}{n}
\DeclareMathAccent{\widecheck}{0}{mathx}{"71}

\def\confName{CVPR}
\def\confYear{2027}

\title{GUITAR-Bench: Benchmarking GUI Agents for Oracle-Free Traversal Testing with Synthesized Defects}

\author{
{\fontsize{11}{14}\selectfont Zixun Wang$^{1}$\quad Xiaolei Li$^{2}$\quad Wenao Ma$^{1}$\quad Mengyang Wu$^{1}$\quad Pingping Zhang$^{3}$\quad Zhijian Hou$^{3,*}$}\\[10pt]
{\fontsize{9}{11}\selectfont
$^{1}$The Chinese University of Hong Kong\quad
$^{2}$Hong Kong University of Science and Technology\quad
$^{3}$City University of Hong Kong}
}
\makeatletter
\newsavebox{\originaltitlebox}
\newlength{\originaltitleheight}
\renewcommand{\maketitle}{%
  \begingroup
  \setbox\originaltitlebox=\vbox{%
    \hsize=\textwidth
    \togglefalse{cvprfinal}
    \@maketitle
  }%
  \setlength{\originaltitleheight}{\dimexpr\ht\originaltitlebox+\dp\originaltitlebox\relax}%
  \twocolumn[\vbox to \originaltitleheight{%
    \hsize=\textwidth
    \centering
    \null\vskip .375in
    {\Large\bfseries\@title\par}%
    \vfill
    {\begin{tabular}{c}\@author\end{tabular}\par}%
    \vfill
  }]%
  \endgroup
}
\def\ps@correspondingfirstpage{%
  \ps@plain
  \def\@oddfoot{%
    \rlap{\normalfont\footnotesize
      \begin{tabular}[t]{@{}l@{}}
        Zixun Wang (\texttt{craddywang@gmail.com}).\\
        $^{*}$Corresponding author: Zhijian Hou (\texttt{zjhou3-c@my.cityu.edu.hk}).
      \end{tabular}%
    }%
    \hfil\normalfont\thepage\hfil
  }%
  \let\@evenfoot\@oddfoot
}
\makeatother
\begin{document}
\maketitle
\thispagestyle{correspondingfirstpage}
\begin{abstract}



As LLM-based coding agents become increasingly capable of generating visual interactive applications,
the bottleneck shifts from code generation to verification. Accordingly, verification moves beyond code-level tests, as many defects surface only through GUI interactions.
This underscores the need for \emph{GUI traversal testing}, which explores applications through their GUIs and reports defective behaviors.
With strong visual reasoning ability, VLM-powered GUI agents offer a promising approach to this task.
However, existing GUI-agent benchmarks evaluate either navigation capability with task instructions or verification capability given specified test oracles, leaving the capabilities required for GUI traversal testing largely unmeasured.
We therefore introduce GUITAR-Bench, a benchmark for evaluating GUI agents on the efficient exploration and oracle construction required by this task.
We propose a novel defect synthesis pipeline designed to satisfy three requirements: realism, low cost, and reliability. The pipeline mines reusable defect patterns from 3k+ real bug fixes to improve realism, automatically synthesizes defects to reduce human effort, and validates the resulting symptom through differential replay to ensure reliability.
The released
benchmark contains 20 web apps and 95 verified GUI defects.
Our evaluation reveals substantial variation among agents using popular models, with recall 
ranging from 6.3\% to 72.6\%. Even the strongest agent, Claude Code backed by Opus 5.5, misses 27.4\% of the defects, showing that traversal testing with GUI agents remains far from solved.

\end{abstract}

\vspace{-1em}
\section{Introduction}
\label{sec:intro}

\begin{figure*}[t]
  \centering
  \vspace{-1em}
  \includegraphics[width=\textwidth]{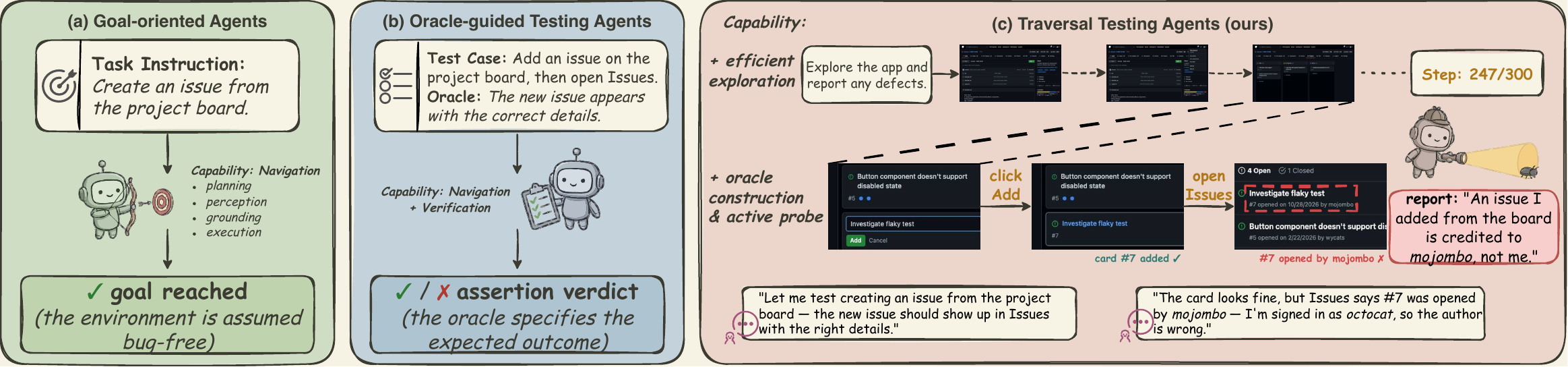}
  \caption{\textbf{Traversal testing versus prior GUI-agent settings.}
  (a)~Goal-oriented agents follow task instructions in a presumed-correct
  environment. (b)~Oracle-guided testers execute human-written checks. (c)~In
  traversal testing, the agent receives only a running app and a generic
  prompt, then explores under budget, constructs oracles, and reports defects.}
  \label{fig:task}
  \vspace{-1.3em}
\end{figure*}

Large language models (LLMs) have revolutionized code generation~\citep{chen2021evaluating,peng2023impact,jimenez2024swe,shi2026swe}. They accelerate modern software development, including visual and interactive
applications such as mobile and web apps.
As the marginal cost of producing code collapses, the economics of application construction invert: the bottleneck shifts from generation to verification~\citep{vaithilingam2022expectation,liu2023your}. 
Accordingly, ensuring application quality now extends beyond traditional unit tests~\citep{jimenez2024swe,chen2021evaluating} to GUI-level testing through end-to-end interactions.

GUI testing dates back decades~\citep{memon2001hierarchical,banerjee2013graphical}, and automated testing tools~\citep{mao2016sapienz,monkey2020} have relied mainly on
random navigation and interaction. These methods lack the semantic understanding of GUI states needed to choose
meaningful actions, and, more importantly, detect only crashes rather than
subtle functional defects.
Vision-language model (VLM)-powered GUI agents can read the rendered screen to act meaningfully and to detect defects beyond crashes, which makes them a
promising approach to GUI testing~\citep{zheng2023judging,zhuge2024agent}.

GUI agents are mainly studied as visual assistants~\citep{zhou2024webarena,xie2024osworld,rawles2025androidworld}, automating users' daily tasks from given instructions (\Cref{fig:task}(a)). This goal-oriented setting emphasizes perception and navigation capabilities and assumes the application itself is bug-free. To adapt GUI agents to application quality assurance, prior work~\citep{hu2024auitestagent,zhang2025artifactsbench,gao2026guitester}, such as WebGen-Bench~\citep{lu2026webgen}, supplies an intent and the expected outcome, and the agent needs verification ability to compare the observed behavior against a known oracle (\Cref{fig:task}(b)). 
In practice, however, test cases and oracles are often unavailable~\citep{barr2014oracle}. This calls for \emph{oracle-free GUI traversal testing} (\Cref{fig:task}(c)), where an agent receives only a running app, explores under an action budget, constructs its own intents and oracles, probes them through interactions, and reports potential defects. 
This setting further requires efficient exploration and oracle construction, forming a progression in capability requirements across the three settings.
Yet whether current agents can meet these demands remains largely unmeasured, motivating a dedicated benchmark.

Such a traversal testing benchmark needs applications whose defects are realistic, cheap to produce, and reliable. 
No existing source satisfies all three:
collections of real bugs are realistic but expensive to reproduce~\citep{just2014defects4j,su2021benchmarking} or annotate~\citep{hong2026framework};
mutation testing is cheap but manufactures faults unlike real
ones~\citep{just2014mutants}; and LLM- or human-injected bugs recover some realism but either lack reliability~\citep{patra2021semantic,ibrahimzada2025challenging} or are costly to create~\citep{teoh2026webtestpilot}.
We satisfy all three desiderata by synthesizing defects from real fixes (see~\cref{fig:pipeline}). Our pipeline first distills 3,595 GUI-related bug fixes into 26 app-agnostic defect patterns for reusable injection into new apps. For each new app, we identify injectable sites in the source code and use an LLM to introduce defects through contextual edits and describe their expected symptom and reproduction scripts. Each script is replayed on clean and defective instances, and the defect is admitted only if the declared symptom appears in the defective one. Thus, distillation from real fixes supplies realism, synthesis keeps cost low, and differential replay ensures reliability. To further mirror the real production environments, defects are sampled to align the distribution with that of real-world bugs.

We release \textbf{GUITAR-Bench} (\textbf{GUI} \textbf{T}esting
\textbf{A}gent on t\textbf{R}aversal testing), a benchmark with synthesized defects for evaluating GUI agents' ability to perform oracle-free traversal testing. It contains 20 web apps and 95 verified GUI defects, including 10 real
open-source projects and 10 mock
apps~\citep{wang2026cua} of commercial products such as Gmail. Each task gives the agent one
app and an action budget, and its reports are scored against
the documented ground truth. We evaluate seven GUI agents on GUITAR-Bench, and find that traversal testing with GUI agents remains far from solved. The strongest agent, Claude Code backed by Opus 5.5, still misses a
substantial fraction of verified defects despite reaching nearly all of
them. Agents often reach the relevant UI but fail to trigger the right behavior,
construct the right oracle, or recognize that the observed outcome is
wrong. Increasing the action budget improves traversal, but does not by
itself remove these post-reach failures. Our contributions are threefold:
\begin{itemize}[leftmargin=*]
    \item We identify the emerging need for oracle-free GUI traversal testing and highlight the need for benchmarks for evaluating GUI agents on demanded capabilities.
    \item We propose a novel defect synthesis approach that generates realistic, low-cost, and reliable defects by synthesizing them from reusable patterns mined from real fixes and validating their symptom through differential replay.
    \item We conduct a diagnostic evaluation of current GUI agents.
    The results show that the main bottleneck is not broader traversal alone, but rather the ability to actively construct oracles and verify outcomes across steps.
\end{itemize}

\vspace{-0.5em}
\section{Related Work}
\label{sec:related}

\subsection{GUI Agents and Their Benchmarks}
GUI agents now operate interfaces as humans do, reading the screen
and acting with clicks and
scrolls~\citep{hong2024cogagent,qin2025ui}. Benchmarks such as
WebArena~\citep{zhou2024webarena}, OSWorld~\citep{xie2024osworld},
and AndroidWorld~\citep{rawles2025androidworld} evaluate these
agents in a goal-oriented setting, where each episode is specified by an explicit task instruction. The agent also assumes the environment
is correct, so when something unexpected happens, it blames its
own action and retries, never suspecting the environment itself. In traversal testing, the agent neither receives any instruction
nor presumes the environment to be bug-free. It must plan its own
exploration, decide what to test and what to
expect, and attribute
unexpected outcomes to its own action or to a defect in the
application. GUITAR-Bench therefore measures the exploration,
defect detection, and attribution abilities that goal-oriented
episodes never exercise.

\subsection{Agent-as-a-Judge for Visual Code Generation}
LLMs increasingly produce software with rendered interface, from design-to-code translation~\citep{si2025design2code,he2026vision2web} to full interactive websites~\citep{lu2026webgen,zhang2025artifactsbench}.
Because correctness often becomes observable only at runtime~\citep{liu2025seeing}, conventional metrics struggle to evaluate these artifacts.
Following the Agent-as-a-Judge
practice~\citep{zheng2023judging,zhuge2024agent}, a growing line of
benchmarks~\citep{lu2026webgen,zhang2025artifactsbench,he2026vision2web,wang2026liveevalbench,yang2026cookiebench,peng2026playcoder} sends GUI agents to judge the generated visual applications.
Throughout this line, the agent serves as the grading tool, yet
its own reliability is unmeasured. GUITAR-Bench is complementary. We turn the judge agent into the subject, casting the judging job as a formal GUI-testing task and measuring the agent's testing ability.

\subsection{GUI Testing and Defect Injection}
GUI testing predates agents by
decades~\citep{memon2001hierarchical,banerjee2013graphical}.
Automated testers such as \textit{Monkey}~\citep{monkey2020} and
\textit{Sapienz}~\citep{mao2016sapienz} perform random
interactions on an app and recognize only crash
bugs~\citep{choudhary2015automated,su2021benchmarking}, leaving
subtler defects to hand-scripted tests. LLM agents
lift these limits, reading the screen to interact meaningfully
and to detect defects beyond
crashes~\citep{yoon2023autonomous,liu2024make}. Most benchmarks for such testing
agents~\citep{hu2024auitestagent,gao2026guitester,zhao2024gui,chen2026guitestscape}
remain oracle-guided, with a human deciding in advance what to
test and what outcome to expect. GUITAR-Bench instead evaluates
traversal testing, where the agent is given no oracle and deciding what to test is an essential part of the task.


CATTest~\citep{hong2026framework} and WebTestPilot~\citep{teoh2026webtestpilot}
are the closest to our oracle-free setting, but they differ in what their defects
are and how their ground truth is established. CATTest uses only AI-generated apps and therefore does not cover failures in real, human-written projects; moreover, its defects require costly manual annotation, as in manually curated benchmarks~\citep{su2021benchmarking}. WebTestPilot covers only four open-source apps and injects one
artificial bug per requirement, with a human-written script as ground truth; thus, its
defects are spec-derived rather than sampled from real fault distributions. In contrast, we use both
maintained projects and AI-generated mocks. We mine 3,595 real fixes into
patterns, inject defects automatically, verify them by differential replay, and sample to
match the real category distribution.

\vspace{-0.2em}
\section{GUITAR-Bench}
\label{sec:method}

\begin{figure*}[t]
  \centering
  \vspace{-1em}
  \includegraphics[width=\textwidth]{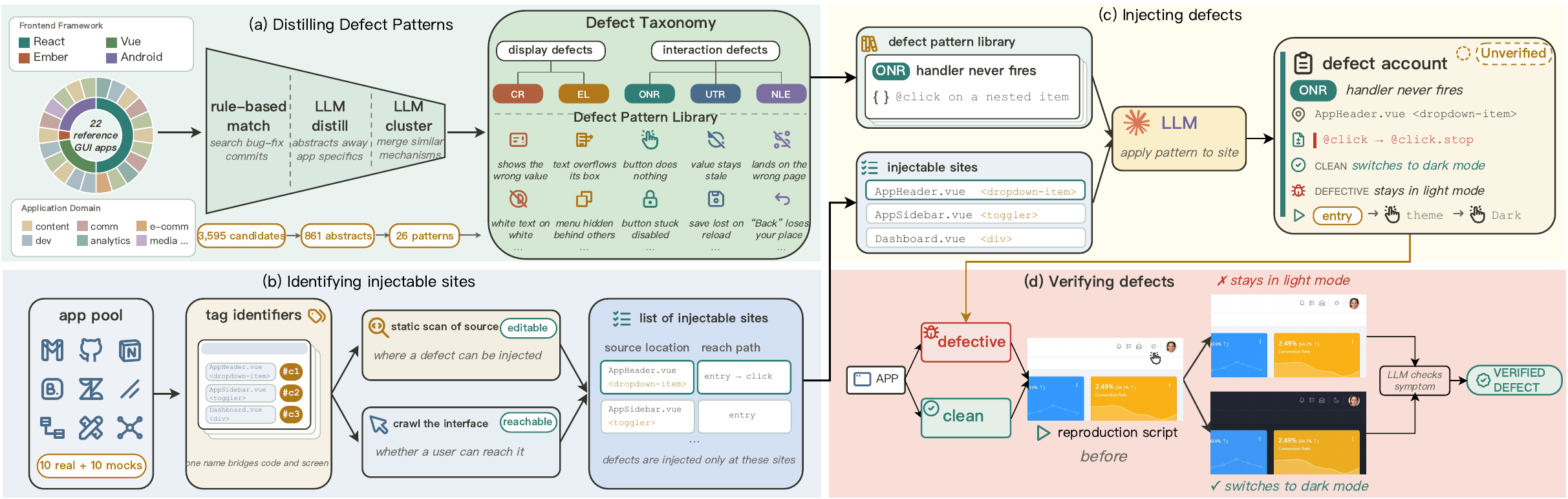}
  \caption{\textbf{The GUITAR-Bench construction pipeline.}
  (a)~We mine real bug-fix commits and distill them into a library of
\emph{defect patterns}.
  (b)~We identify each application's \emph{injectable sites}, code
locations that render a UI element reachable in the running
interface.
  (c)~An LLM matches a pattern to a site and writes the edit, along with a \emph{defect account} that declares the correct behavior before injection, the expected symptom after it, and a script to reproduce it.
  (d)~The script is replayed on a clean and a defective instance, and an LLM compares the two runs against the account. The defect is admitted only if the declared symptom appears on the defective instance.}
  \label{fig:pipeline}
  \vspace{-1.3em}
\end{figure*}


\subsection{Task Formulation}
\label{sec:task}

A traversal-testing task is described as a triple $(\mathcal{A}, \mathcal{E}, B)$: a
running web app $\mathcal{A}$, a set $\mathcal{E}$ of GUI defects
present in it, and an interaction budget of $B$ steps. The agent is given
$\mathcal{A}$ alone, while $\mathcal{E}$ is held by the evaluator. The episode
is a finite-horizon POMDP: the underlying state comprises
the full frontend and backend state of $\mathcal{A}$, of which the agent
observes only a screenshot $o_t$ of the current page at step $t$. Its action space is the standard GUI-agent vocabulary of
coordinate-based pointer and keyboard operations (e.g., \texttt{click},
\texttt{type}, \texttt{scroll}, plus browser-level \texttt{back} and
\texttt{reload})~\citep{zhou2024webarena,xie2024osworld}, extended with
one task-specific tool, $\texttt{report(description, location)}$, which
files a natural-language statement of what is wrong and where, and
leaves the environment state unchanged. 
The only guidance is a generic system prompt that instructs the agent to traverse the interface and
report the defects it finds. Beyond this prompt, the episode carries no
task-specific instruction, no test case, and no success signal. The episode
terminates when the budget is exhausted or when exploration converges.
The agent's output is the report set $\mathcal{R}$ accumulated over
the episode, together with the trajectory that produced it, and is
scored against $\mathcal{E}$ once the episode ends.

\subsection{Benchmark Overview}
\label{sec:overview}

GUITAR-Bench consists of tasks
$\{(\mathcal{A}_i, \mathcal{E}_i, B)\}_{i=1}^{T}$, where
$\mathcal{A}_i$ is an application seeded with the verified
defect set $\mathcal{E}_i$. The 20 web apps span communication,
e-commerce, software development, and enterprise management
~\citep{zhou2024webarena,xie2024osworld,rawles2025androidworld}.
Ten are real open-source projects from GitHub, reflecting the conventions of human-written code. The other ten are mock
apps from CUA-Gym-Hub~\citep{wang2026cua} that reproduce
the core frontend of widely used commercial products such as Gmail
and Notion. 
These products are closed-source, so the mocks are
the only way to bring their interfaces, and
they remain fully editable and servable by the pipeline. They are,
moreover, themselves AI-generated software, exactly the class of
application whose verification motivates this work. In principle,
the pipeline generalizes to any web application whose source code
is available; we restrict the benchmark to these 20 to keep it
reproducible and tractable.

An automated pipeline synthesizes each task instance in two stages. In
the first stage, performed once for all applications, an LLM reads a
corpus of real bug-fix commits from open-source GUI projects and
distills them into a library of \emph{defect patterns}, recurring
buggy code edits paired with the visible symptoms they produce
(Sec.~\ref{sec:patterns}). In the second stage, starting from a clean
instance of each application, an LLM turns a pattern into an edit
of the source code, declaring its expected symptom, a \emph{reproduction script} that should trigger it, and when the symptom becomes observable.
Each declaration is then verified by replaying its reproduction script on
clean and defective instances and using an LLM to check whether the
declared symptom appears only after injection.
Admitted defects form a pool from which the released set is
composed to a target distribution (Sec.~\ref{sec:distribution}). \Cref{fig:pipeline} illustrates the two-stage pipeline, with pattern distillation in (a) and defect injection in (b--d).

\subsection{Distilling Defect Patterns from Real Bug Fixes}
\label{sec:patterns}
Defects generated purely by an LLM would inherit the model's training-data biases and may not reflect how real GUI software fails. We therefore mine defect patterns from real GitHub bug fixes, drawing on 22 open-source applications that span three web
frameworks (React, Vue, and Ember) and Android. 
They cover prior GUI-agent benchmark domains, including e-commerce,
content management, analytics dashboards, and team communication
~\citep{zhou2024webarena,xie2024osworld,rawles2025androidworld}.
Within each repository we select, through rule-based git and regular-expression matching, the bug-fix commits whose effect is visible in the GUI. A fix enters the corpus through one of two signals. Interaction bugs usually ship a regression test that exercises the GUI rather
than a bare function, which we keep as the bug's reproduction.
Display bugs rarely do, so we admit them by commit messages
reporting a rendering, layout, or styling fix and take the change
from the diff. The selection yields 3{,}595 candidate defects.

An LLM then abstracts each candidate into an application-agnostic pattern: the expected behavior it breaks, the interaction that exposes it, and a \emph{code signature}, the structural code pattern that locates injectable sites in a new application. To keep the library from being dominated by the few largest repositories, we sample at most 45 candidates per repository, stratified by the length of their triggering sequence. In the end, we keep the 26 patterns whose effect surfaces in the interface, the ones a defect injector can reproduce and a GUI agent can observe.

The patterns are then organized by defect category. The categories are fixed: the five defined by GTArena~\citep{zhao2024gui} and adopted by recent GUI-testing benchmarks~\citep{gao2026guitester,chen2026guitestscape}, namely content-rendering (CR) and element-layout (EL) display defects, and operation-no-response (ONR), unexpected-task-result (UTR), and navigation-logic-error (NLE) interaction defects. Beneath each category sit its distilled patterns, 26 in total, which form the defect pattern library. When injecting defects, we follow the mined frequencies over categories, so the benchmark's defect distribution mirrors that of real-world bugs (Sec.~\ref{sec:distribution}).

\subsection{Synthesizing Defects with Language Models}
\label{sec:defects}

\tightpara{Identifying injectable sites} A \emph{site} is a source code location that could host a
defect edit. The edit's symptom, however, must surface in the
interface, so a site is \emph{injectable} only if it renders a UI
element that a user can reach. To establish this
correspondence, a static scan first collects candidate sites and
tags each with a stable identifier. When the application runs, a
site's identifier surfaces in the Document Object Model (DOM) on
the element it renders, so matching identifiers ties every site to
its UI element, or reveals that it renders none. A crawler then
explores the running application, following the identifiers to
record an action path from the entry page to each matched element.
Sites whose element is reached in this way form the
application's \emph{injectable sites}, the unit of injection for
all later stages.

Each mined pattern carries a \emph{code signature}, which records the code
structure the pattern needs in a new application, such as an event handler
or a navigation call. A pattern is feasible at a site when the site's source file contains at
least one of them. The filter is deliberately loose, and the LLM may
still decline a pair it cannot host.

\tightpara{Injecting defects into source code}
Given an application's injectable sites and the pattern library, we pair the two before any edit is written. The pipeline samples injections from the feasible \emph{(site, pattern)} pairs following real distribution (Sec.~\ref{sec:distribution}). For each sampled pair, an LLM acting
as a developer introduces the bug with a small, contextual edit in
the style of the surrounding code. The edit stays local to its
element, and the application still builds and renders normally, so
the defect reveals itself only through traversal. A display defect
requires reaching the affected screen, and an interaction defect
additionally requires exercising the behavior. Alongside the edit, the model declares an account of the defect: the correct behavior before injection, the symptom a user would observe, and a \emph{reproduction script} (a short, deterministic interaction sequence that should make the defect appear, with navigation taken from the recorded action paths). At this point the account is a
set of claims rather than facts. The next stage tests every claim by
execution, and the verified account then serves as the ground truth
against which testing-agent reports are judged, making
GUITAR-Bench verifiable by construction.

\tightpara{Verifying defects by execution}
One admission rule governs this stage: \emph{no defect enters
GUITAR-Bench until it is verified, by execution, to visibly appear in
the interface}. We boot the clean and the defective instance from the
same seed state and replay the reproduction script on both. When a
scripted step no longer matches the page, an LLM agent repairs that
single step from the current screenshot rather than abandoning the
replay. Once both instances reach the observation screen, an LLM then compares their screenshots and admits the defect
only if the declared symptom is actually visible on the defective side. The admitted account is stored as the defect's ground truth for judging an agent's reports.

\vspace{-0.7em}
\section{Benchmark Statistics}
\label{sec:distribution}

This section characterizes the defects in GUITAR-Bench. We first
describe them along three axes (Sec.~\ref{sec:design_dist}) and then
show how the released set is shaped along these axes
(Sec.~\ref{sec:align_dist}). Finally, we report the results of a human
quality check, namely the agreement and pass rate
(Sec.~\ref{sec:qc}).

\vspace{-0.3em}
\subsection{Designing the Distribution Axes}
\label{sec:design_dist}

Let $S_{\mathcal{A}}$ be the injectable sites of an application $\mathcal{A}$ and $\mathcal{P}$ the distilled defect patterns. 
A candidate injection is a pair $(s,p)\in S_{\mathcal{A}}\times \mathcal{P}$. Not every pair is
compatible. A pattern applies only where its code signature matches the site's structure,
so we keep the feasible set
\vspace{-0.6em}
\begin{equation*}
F_{\mathcal{A}}=\{(s,p)\in S_{\mathcal{A}}\times \mathcal{P} :
\text{$p$'s code signature matches $s$}\},
\end{equation*}
decided by a lightweight static filter. Each feasible pair is characterized along three axes. The pattern
carries the defect category
$\kappa(p)\in\{\mathrm{CR},\mathrm{EL},\mathrm{ONR},\mathrm{UTR},\mathrm{NLE}\}$
and a \emph{symptom distance} $\delta(p)\in\{\mathrm{near},\mathrm{far}\}$.
A near defect shows on the very screen where it is triggered,
while a far defect looks fine there and surfaces only elsewhere,
for example on another page reached by further navigation. Display
defects admit only $\delta=\mathrm{near}$, since their symptom is
rendered the moment the element is. For an interaction defect, the
distance is first estimated during distillation from how the
pattern's mined bugs surfaced, which serves as a prior for
sampling, and settled only at verification, where a defect is
labeled far if the clean and defective instances look the same
right after the trigger and differ only once the screen is left.

The site determines the navigation depth
$\rho(s)\in\{\mathrm{shallow},\mathrm{deep}\}$, deep when reaching it takes at least three interactions from the
entry page, and shallow otherwise. The threshold follows the
three-click heuristic of interface design, which holds that any function
should be reachable within about three clicks of the landing
page~\citep{zeldman2001talent}. It also matches our applications, where sites at depth three or more
lie beyond the 85th percentile of all injectable sites.

The three axes partition the feasible pairs into cells
$(\kappa,\rho,\delta)$. Injected defects then induce a
joint distribution over the cells, $\pi_{\kappa,\rho,\delta}$
being the share of defects that simultaneously fall in category
$\kappa$, at depth $\rho$, and at distance $\delta$. We do not prescribe the joint but only its marginals
$\pi_\kappa$, $\pi_\rho$, and $\pi_\delta$, leaving the joint free
to accommodate structural constraints, such as display defects
admitting no far symptom.

\subsection{Aligning the Distribution}
\label{sec:align_dist}

\begin{figure}[t]
  \centering
  \includegraphics[width=\columnwidth]{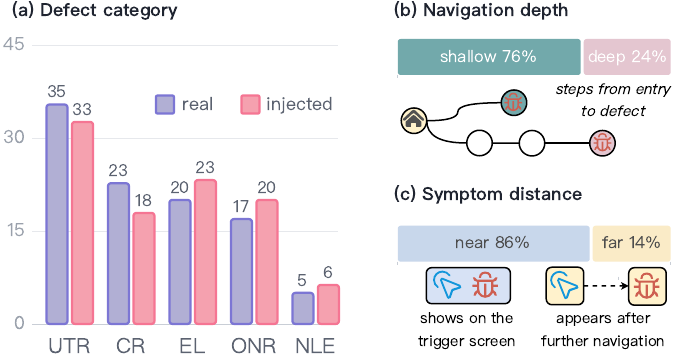}
  \caption{\textbf{Distribution of GUITAR-Bench.} (a)~Defect categories of the released benchmark (injected) closely follow those of real bug fixes mined from open-source projects (real).
  (b)~Navigation depth counts the interactions from the entry page
  until the defective element becomes visible. A defect is deep if reaching it takes at least three interactions from the entry page; 24\% of the defects are deep.
  (c)~Symptom distance is how far from the triggering action the defect becomes visible: a near defect shows on the same screen, while a far
  one appears only after leaving it; 14\% are far.}
  \label{fig:distribution}
\end{figure}

We steer the marginals in two ways. The category marginal $\pi_\kappa$
follows the frequencies of the mined corpus, so the benchmark reflects
which failures actually occur in real software. Depth and distance, in
contrast, are bounded by supply. Deep sites are rare in real
applications, and only some patterns can produce a far symptom. Within
these limits, we sample as many deep and far defects as possible.

Sampling once does not reach this goal, because verification discards
defects unevenly across cells. A defect intended to be far, for
instance, often turns out to surface on the spot. We therefore inject
in rounds. Each round gives every cell a number of slots proportional
to $\pi_\kappa\pi_\rho\pi_\delta$, with the depth and distance shares
set above what the supply allows. The next round then refills only the
cells that fall short after verification. The released set is chosen
from the verified defects by the same rule, spread evenly across
applications, composed into one instance per application, and reviewed
by two annotators (Sec.~\ref{sec:qc}).

Figure~\ref{fig:distribution} summarizes the released benchmark, which
contains $N=95$ verified defects across the 20 applications. Category
shares stay within five points of the mined frequencies for every class.
Of all defects, 24\% are deep, and
14\% of the interaction defects are far.

\subsection{Human Quality Check}
\label{sec:qc}

The automatic verification filters out most faulty injections, but its
final verdict still rests on an LLM, so some defects may not behave as
intended. To check every defect by hand, we divide the 20 applications
into six groups, and six annotators each take two groups, so that
every group is reviewed by two people independently. A reviewer
follows the reproduction steps of each defect and judges whether it
can be triggered and whether the described symptom appears. The two
reviewers agree on 85\% of the defects. For the rest, an additional annotator
collects both opinions and re-checks the defect to reach a verdict.
In the end, 95 of the 103 checked defects (92\%) pass and form the
final benchmark, which indicates that the construction pipeline is
reliable.

\vspace{-0.5em}
\section{Experiments}
\label{sec:exp}

\begin{table*}[t]
  \centering
  \small
  \setlength{\tabcolsep}{4pt}
  \caption{\textbf{Main results on GUITAR-Bench.} Cov., Reach, and Rec.$|$R. denote Coverage, Reach, and Recall among reached defects. Category, depth, and
  distance columns report recall on each subset. Bold marks the best
  model within the simple harness.}
  \label{tab:main}
  \begin{tabular}{l cccc ccccc cc cc}
    \toprule
    & \multicolumn{4}{c}{Overall} & \multicolumn{5}{c}{Category} & \multicolumn{2}{c}{Depth} & \multicolumn{2}{c}{Distance} \\
    \cmidrule(lr){2-5} \cmidrule(lr){6-10} \cmidrule(lr){11-12} \cmidrule(lr){13-14}
    Model & Cov. & Reach & Recall & Rec.$|$R. & EL & CR & ONR & UTR & NLE & Shallow & Deep & Near & Far \\
    \midrule
    \multicolumn{14}{l}{\textit{Claude Code}} \\
    \quad Opus 5.5 & 80.0 & 94.7 & 72.6 & 76.7 & 90.9 & 76.5 & 63.2 & 64.5 & 66.7 & 72.2 & 73.9 & 74.7 & 50.0 \\
    \midrule
    \multicolumn{14}{l}{\textit{Simple harness (ours)}} \\
    \quad Opus 5.5 & 68.6 & 83.2 & \textbf{42.1} & \textbf{49.4} & \textbf{45.5} & \textbf{52.9} & \textbf{57.9} & 25.8 & 33.3 & \textbf{40.3} & \textbf{47.8} & \textbf{43.7} & \textbf{25.0} \\
    \quad DeepSeek-V4-Pro & 65.1 & 82.1 & 27.4 & 33.3 & 22.7 & 23.5 & 26.3 & \textbf{29.0} & \textbf{50.0} & 27.8 & 26.1 & 28.7 & 12.5 \\
    \quad Gemini 3.8 Flash & \textbf{74.0} & \textbf{87.4} & 25.3 & 28.9 & 27.3 & 29.4 & 36.8 & 16.1 & 16.7 & 25.0 & 26.1 & 26.4 & 12.5 \\
    \quad GLM-5.3-Flash & 61.3 & 78.9 & 18.9 & 24.0 & 22.7 & 5.9 & 10.5 & 25.8 & 33.3 & 19.4 & 17.4 & 18.4 & \textbf{25.0} \\
    \quad Qwen3.7-Plus & 52.8 & 74.7 & 13.7 & 18.3 & 18.2 & 5.9 & 26.3 & 9.7 & 0.0 & 12.5 & 17.4 & 14.9 & 0.0 \\
    \quad Doubao-Seed-2.1-Turbo & 58.9 & 71.6 & 6.3 & 8.8 & 4.5 & 5.9 & 10.5 & 3.2 & 16.7 & 6.9 & 4.3 & 6.9 & 0.0 \\
    \bottomrule
  \end{tabular}
\end{table*}

We evaluate GUI testing agents on GUITAR-Bench through four research
questions:
\begin{itemize}[leftmargin=*]
    \item \textbf{RQ1.} How effective are current visual agents at
    traversal testing?
    \item \textbf{RQ2.} How do defect category, depth, and symptom distance affect agent performance?
    \item \textbf{RQ3.} How does the action budget affect detection?
    \item \textbf{RQ4.} What common failure patterns do agents exhibit?
\end{itemize}

\vspace{-0.15em}
\subsection{Experimental Setup}
\vspace{-0.3em}
\label{sec:setup}

\tightpara{Agents}
All agents interact with the benchmark through screenshots and
pixel-coordinate actions, with no access to the DOM or accessibility tree.
This visual-only setup is more general because it does not assume
structured page state is available. As a
strong commercial-agent baseline, we run Claude Code backed by Opus
5.5~\citep{anthropic2026claudecode,anthropic2026opus55}, which includes its own planning and memory mechanisms. To
compare foundation models under a controlled interface, we also implement
a lightweight ReAct-style harness~\citep{yao2023react} that provides the agent with its three most recent
screenshots, persistent notes, and the pages visited so far. We run this harness with Opus 5.5, Gemini 3.8 Flash~\citep{google2026gemini38flash},
DeepSeek-V4-Pro~\citep{xu2026deepseek},
Qwen3.7-Plus~\citep{qwen2026qwen37plus}, GLM-5.3-Flash~\citep{zai2026glm53flash}, and Doubao-Seed-2.1-Turbo~\citep{bytedance2026seed21}. Running Opus 5.5 in both harnesses separates the effect
of the harness from that of the model. We empirically set the action budget to $B=300$ for the main experiments
and study the effect of the budget in RQ3. We report full model identifiers in tables and use shortened names in the
text for readability.

\tightpara{Evaluation Metrics}
Our primary metric is \emph{Recall}, the fraction of injected defects
matched by an agent report. A defect is \emph{reached} if any trigger site
appears on screen. We report \emph{Reach} and \emph{Rec.$|$R.}, recall
among reached defects, to identify where defects are lost.
Recall is also broken down by category, depth, and distance.
We defer precision analysis to the
appendix because the applications may contain pre-existing bugs outside
our injected set; therefore, a report that does not match an injected
defect is not necessarily false. Beyond defect detection, we measure
\emph{Coverage} (\emph{Cov.}), defined as the fraction of UI sites that
ever appear on the agent's screen, to quantify exploration efficiency.

\begin{figure}[t]
  \centering
  \includegraphics[width=\columnwidth]{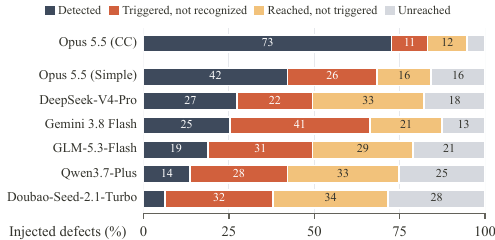}
  \caption{\textbf{Detection progress decomposition.} Each bar splits
  injected defects in the detection process: unreached, reached but not triggered, triggered but not recognized, or detected.}
  \label{fig:funnel}
\end{figure}

\vspace{-0.4em}
\subsection{RQ1: Overall Effectiveness}
\label{sec:rq1}

\Cref{tab:main} presents the overall results and highlights two main findings. First, GUITAR-Bench separates agents widely, with Recall ranging from 6.3\% to 72.6\%. Second, the benchmark is far from saturated. Even the strongest agent, Claude Code backed by Opus, misses 27.4\% of the injected defects despite reaching nearly all of them, showing that GUI traversal testing remains an open challenge.

Both the model and the harness matter. In the simple harness, Opus
leads with 42.1\% Recall, and the other models range from 6.3\% to
27.4\%. Running Opus with Claude Code raises Recall from 42.1\% to 72.6\%, a gap twice as large as that between the two best models within the simple harness. Rec.$|$R. increases more than Reach through this change, indicating that the stronger harness mainly helps convert reached defects into reports. 

This discrepancy indicates that exploration does not necessarily lead to successful testing. For example, Qwen and Doubao achieve comparable Coverage, but Qwen attains twice the Recall. Similarly, Gemini achieves higher Coverage yet detects fewer defects than DeepSeek. RQ2 examines what happens between reaching and reporting a defect.

\subsection{RQ2: Performance Across Defect Attributes}
\label{sec:rq2}
\begin{table*}[!b]
  \centering
  \small
  \setlength{\tabcolsep}{4pt}
  \caption{\textbf{Representative defect patterns and missed behaviors.}
  Recall is averaged over the six simple-harness models. \emph{Untrig.} is
  the share of reached-but-missed runs in which the defect was never
  triggered.}
  \label{tab:patterns}
  \renewcommand{\arraystretch}{1}
  \begin{tabular}{>{\raggedright\arraybackslash}p{2.7cm} c c >{\raggedright\arraybackslash}p{5.6cm} >{\raggedright\arraybackslash}p{5.8cm}}
    \toprule
    Pattern & Recall & Untrig. & Example defect & Typical failure \\
    \midrule
    Wrong destination (NLE)
      & 38.9 & 60\%
      & Clicking a contact in the inbox opens the chat of the next person
        in the list.
      & The agent clicks ``Alex John'', lands in Taylor's chat, and types
        ``Hello Taylor!''. \\
    Detached handler (ONR)
      & 37.9 & 73\%
      & The \emph{Features} link in the top bar should jump down the page
        but does nothing.
      & The agent never clicks the link and heads into the editor or
        scrolls by hand. \\
    Incorrect computation (UTR)
      & 25.6 & 51\%
      & A Roman numeral converter turns 42 into XLI and 421 into CDXX.
      & The agent converts 421, gets CDXX, notes that it ``works
        normally'', and moves on. \\
    Stale or unpersisted state (UTR)
      & 8.3 & 60\%
      & After an item is created, the form resets but the dialog never closes.
      & The agent sees the dialog stay open and assumes it is meant for
        batch creation. \\
    Incorrect enabled state (ONR)
      & 0.0 & 92\%
      & The \emph{Create} button turns grey once anything is typed into
        the search box.
      & The agent never clicks \emph{Create} while a search is active,
        the only trigger. \\
    \bottomrule
  \end{tabular}
  \vspace{-0.6em}
\end{table*}

We first relate Recall to the three attributes defined in~\cref{sec:design_dist},
and then decompose where defects are lost.

\subsubsection{Defect Attributes}

\tightpara{Defect category}
The category columns of~\cref{tab:main} show that display defects are
generally easier than interaction defects. EL and CR defects can be
noticed once the affected element is rendered, whereas ONR, UTR, and NLE
require the agent to trigger the element and judge the outcome. For weaker
agents, even display defects are often missed, showing that
screenshot-level anomaly recognition remains challenging.
The interaction
categories further show that models have different failure profiles, not
just different overall Recall. Gemini and DeepSeek have similar overall
Recall (25.3\% vs.\ 27.4\%), but Gemini is stronger on ONR while
DeepSeek is stronger on UTR and NLE.

\tightpara{Navigation depth} Perhaps surprisingly, depth only weakly correlates with difficulty.
Claude Code recalls 73.9\% of deep defects and 72.2\% of shallow ones,
and other agents show similar gaps. Depth
does make defects harder to reach, but this does not translate into a
large Recall gap. This suggests that many agents can traverse far enough
to expose deep UI regions, while the harder part is still deciding
whether the observed behavior is wrong.

\tightpara{Symptom distance}
Far defects are the hardest for every agent. Claude Code detects 50.0\%
of far defects, compared with 74.7\% of near ones, and the best
simple-harness models detect only 25.0\%. For far defects, the symptom is
not observed at the triggering action but appears only in a later state.
The agent must connect that later observation back to an earlier
action. Far defects thus capture a central difficulty of traversal testing:
the agent must not only visit the right UI, but also maintain an oracle
across actions.

\subsubsection{Detection Progress}

The above analysis shows that harder defects require triggering and checking, but Reach and Recall do not reveal where agents lose them. We therefore decompose detection progress into four statuses. \Cref{fig:funnel} shows the resulting breakdown.
A defect is
\emph{triggered} once the agent performs the interaction that activates
it, or, for a display defect, once the affected element is rendered in its
defective state. A triggered defect is detected only if the agent
recognizes the symptom and reports it. 

The breakdown shows that post-reach failures are heterogeneous. Using the same model, Claude Code and the simple harness trigger similar shares of reached defects (87.8\% vs. 81.2\%), but Claude Code recognizes substantially more triggered defects (87.3\% vs. 61.5\%). This suggests that the harness primarily improves recognition rather than triggering. Across models, both triggering and recognition capabilities vary. Gemini triggers more reached defects than
DeepSeek but recognizes fewer, which partially explains why DeepSeek has higher UTR
Recall despite lower Coverage. Qwen and Doubao reach and trigger similar
shares of defects, but Qwen converts more of them into reports.

\subsection{RQ3: Effect of Action Budget}
\label{sec:rq3}

\begin{figure}[t]
  \centering
  \begin{subfigure}{0.49\columnwidth}
    \centering
    \includegraphics[width=\linewidth]{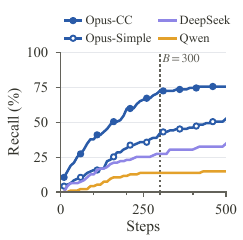}
    \caption{Recall}
    \label{fig:budget_recall}
  \end{subfigure}\hfill
  \begin{subfigure}{0.49\columnwidth}
    \centering
    \includegraphics[width=\linewidth]{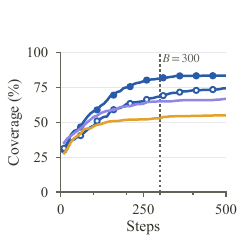}
    \caption{Coverage}
    \label{fig:budget_coverage}
  \end{subfigure}
  \caption{\textbf{Effect of the action budget.} Recall and Coverage for
  representative agents, evaluated at increasing action cutoffs.}
  \label{fig:budget}
\end{figure}

To study whether larger budgets close the problem, we run agents for up to 500
actions and evaluate smaller budgets by truncating trajectories.
~\Cref{fig:budget} shows that increasing budgets improves traversal but with diminishing returns.
Coverage rises quickly in the first 150 actions, then slows and converges
to similar levels across agents, while Recall grows more gradually and remains
widely separated. Claude Code reaches 37.9\% Recall within 100 actions,
nearly matching simple-harness Opus at 300 actions (42.1\%), while other
agents remain much lower. Thus, more actions help exploration, but the
defects found per action still depend strongly on the model and harness.

The gains also taper unevenly. Claude Code improves by only 3.2 points
beyond $B=300$, whereas simple-harness Opus
still gains 10.5 points. We use $B=300$ as a cost-effective main budget.
It captures the near-saturated regime for the strongest system, while this
analysis quantifies the additional benefit of longer runs. Consistent with RQ2,
longer runs can help agents reach and trigger more defects, but they do
not by themselves solve the post-reach bottleneck of recognizing that the
observed behavior is wrong.

\subsection{RQ4: Common Failure Patterns}
\label{sec:rq4}

\begin{figure}[t]
  \centering
  \vspace{-0.6em}
  \includegraphics[width=\columnwidth]{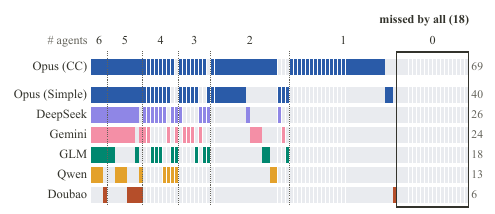}
  \caption{\textbf{Agents miss the same defects.} Each column is one
  injected defect, sorted by how many of the seven agents detect it at
  $B=300$. Colored cells mark detections and gray cells mark misses. The
  boxed block contains the defects that no agent detects.}
  \label{fig:overlap}
  \vspace{-0.3em}
\end{figure}

\tightpara{Failure overlap}
\Cref{fig:overlap} shows highly nested detections. Most defects found by a
simple-harness model are also found by Claude Code, yet 18 defects are
missed by every agent. Pooling all six
simple-harness models detects 53 defects, still below Claude Code alone,
suggesting shared difficulty rather than complementary weaknesses.

\tightpara{Pattern-level causes}
The overlap analysis shows a shared hard set, but individual
defects are too sparse to explain why. We therefore analyze predefined
patterns from~\cref{sec:patterns}, which link defects to the testing behaviors needed to expose them. \Cref{tab:patterns} shows representative
patterns across interaction categories. Extending RQ2's category analysis,
the pattern view shows difficulty also varies within a category.
Detached handlers are detected 37.9\% of the time, while incorrect enabled
states are never detected by any simple-harness model. They are hard to
trigger because the agent must perform a sequence, such as typing in the
search box and then checking whether \emph{Create} is still usable. A
detached handler, in contrast, needs only a direct click. These
patterns expose both triggering and recognition bottlenecks. Closing these
gaps requires traversal-testing capabilities: forming test intents,
constructing oracles, and verifying outcomes across multiple steps.

\subsection{Closing the Loop with Repair}
\label{sec:repair}

Testing matters only if its reports can drive repair. GUITAR-Bench lets
us evaluate this test-to-fix loop without additional human labeling:
each defect is synthesized with a complete account, and replaying
its reproduction steps on the clean app provides the reference behavior
for checking repairs. We run one
loop iteration with Claude Code backed by Opus 5.5 as both tester and
developer. After testing, reports matched to defects are given to
the developer, which edits the app based on the reported location and
symptom. A patch counts as fixed if it exactly reverts the injected edit
or makes the app match the reference behavior.

\Cref{tab:repair} shows that detected defects are usually actionable:
67 of 69 detected defects are successfully repaired (97.1\%). The two
unrepaired cases are not coding failures: both address the reported issue
but differ from the clean app in arguably valid ways. For
example, when told that a \emph{Reserve} button did nothing, the developer
made it scroll to the room list, whereas the clean app opened checkout.

\subsection{Reliability of the Report Judge}
\label{sec:judge_audit}

\begin{table}[t]
  \centering
  \small
  \setlength{\tabcolsep}{3.5pt}
  \caption{\textbf{Test-to-fix.} Most of the detected defects are repaired.}
  \label{tab:repair}
  \begin{tabular}{lccc}
    \toprule
    Defect type & Total & Detected & Fixed / detected \\
    \midrule
    All & 95 & 69 (72.6\%) & 67/69 (97.1\%) \\
    Display & 39 & 33 (84.6\%) & 33/33 (100\%) \\
    Interaction & 56 & 36 (64.3\%) & 34/36 (94.4\%) \\
    \bottomrule
  \end{tabular}
\end{table}

Our recall metric uses an LLM-as-a-judge to match free-form reports to defect descriptions. We audit the judge on a stratified sample of 60 decisions: 25 matched pairs, 15 near-miss negatives, 10 random negatives, and 10 far-defect cases. Two annotators independently inspect each case and then resolve disagreements for a consensus. \Cref{tab:judge_agreement} shows high inter-annotator agreement, and the judge agrees with the consensus on 96.7\% of decisions
($\kappa=0.93$), indicating that it is reliable for evaluating testing agents.

\begin{table}[H]
  \centering
  \small
  \setlength{\tabcolsep}{4pt}
  \caption{\textbf{Judge audit agreement.} 
  \emph{Judge--human} compares the judge with the consensus of two
annotators, while \emph{human--human} compares the two annotators. $\kappa$ is
  Cohen's $\kappa$.}
  \label{tab:judge_agreement}
  \begin{tabular}{lcc}
    \toprule
    & Judge--human & Human--human \\
    \midrule
    Agreement ($\kappa$) & 96.7\% (0.93) & 93.3\% (0.87) \\
    \bottomrule
  \end{tabular}
\end{table}
\section{Conclusion}
\label{sec:conclusion}
We introduced GUITAR-Bench, a benchmark for evaluating GUI agents on oracle-free GUI traversal testing, comprising 20 applications and 95 synthesized defects. Our evaluation shows that current visual agents remain far from
reliable GUI testers: even after reaching the relevant UI, they often fail
to trigger the right behavior, construct the right oracle, or recognize
faulty outcomes. Together with the repair-loop result, these findings call for more
attention to oracle-free traversal testing.

\par\penalty-10000
\clearpage
{\small
\bibliographystyle{ieeenat_fullname}
\bibliography{references}}

@inproceedings{jimenez2024swe,
  title={Swe-bench: Can language models resolve real-world github issues?},
  author={Jimenez, Carlos E and Yang, John and Wettig, Alexander and Yao, Shunyu and Pei, Kexin and Press, Ofir and Narasimhan, Karthik},
  booktitle={International Conference on Learning Representations},
  volume={2024},
  pages={54107--54157},
  year={2024}
}

@article{shi2026swe,
  title={SWE-Bench ProMax: Benchmarking Agents on Large-Scale Multilingual Code Refactoring},
  author={Shi, Yuling and Xu, Jinghan and Fu, Kelin and Zeng, Wenhao and He, Shilin and Zhang, Lei and Liu, Yue and Zhao, Zelin and Zhuo, Terry Yue and Cao, Jialun and others},
  journal={arXiv preprint arXiv:2608.09802},
  year={2026}
}

@article{lu2026webgen,
  title={Webgen-bench: Evaluating llms on generating interactive and functional websites from scratch},
  author={Lu, Zimu and Yang, Yunqiao and Ren, Houxing and Hou, Haotian and Xiao, Han and Wang, Ke and Shi, Weikang and Zhou, Aojun and Zhan, Mingjie and Li, Hongsheng},
  journal={Advances in Neural Information Processing Systems},
  volume={38},
  year={2026}
}

@article{chen2021evaluating,
  title={Evaluating large language models trained on code},
  author={Chen, Mark and Tworek, Jerry and Jun, Heewoo and Yuan, Qiming and Pinto, Henrique Ponde De Oliveira and Kaplan, Jared and Edwards, Harri and Burda, Yuri and Joseph, Nicholas and Brockman, Greg and others},
  journal={arXiv preprint arXiv:2107.03374},
  year={2021}
}

@article{zhang2025artifactsbench,
  title={Artifactsbench: Bridging the visual-interactive gap in llm code generation evaluation},
  author={Zhang, Chenchen and Li, Yuhang and Xu, Can and Liu, Jiaheng and Liu, Ao and Zhou, Changzhi and Deng, Ken and Wu, Dengpeng and Huang, Guanhua and Li, Kejiao and others},
  journal={arXiv preprint arXiv:2507.04952},
  year={2025}
}

@article{he2026vision2web,
  title={Vision2web: A hierarchical benchmark for visual website development with agent verification},
  author={He, Zehai and Hong, Wenyi and Yang, Zhen and Pan, Ziyang and Liu, Mingdao and Gu, Xiaotao and Tang, Jie},
  journal={arXiv preprint arXiv:2603.26648},
  year={2026}
}

@inproceedings{gao2026guitester,
  title={Guitester: Enabling gui agents for exploratory defect discovery},
  author={Gao, Yifei and Wu, Jiang and Chen, Xiaoyi and Yang, Yifan and Cui, Zhe and Ma, Tianyi and Zhang, Jiaming and Sang, Jitao},
  booktitle={Findings of the Association for Computational Linguistics: ACL 2026},
  pages={18956--18978},
  year={2026}
}

@inproceedings{vaithilingam2022expectation,
  title={Expectation vs. experience: Evaluating the usability of code generation tools powered by large language models},
  author={Vaithilingam, Priyan and Zhang, Tianyi and Glassman, Elena L},
  booktitle={Chi conference on human factors in computing systems extended abstracts},
  pages={1--7},
  year={2022}
}

@article{liu2023your,
  title={Is your code generated by chatgpt really correct? rigorous evaluation of large language models for code generation},
  author={Liu, Jiawei and Xia, Chunqiu Steven and Wang, Yuyao and Zhang, Lingming},
  journal={Advances in neural information processing systems},
  volume={36},
  pages={21558--21572},
  year={2023}
}

@article{zhao2024gui,
  title={Gui testing arena: A unified benchmark for advancing autonomous gui testing agent},
  author={Zhao, Kangjia and Song, Jiahui and Sha, Leigang and Shen, Haozhan and Chen, Zhi and Zhao, Tiancheng and Liang, Xiubo and Yin, Jianwei},
  journal={arXiv preprint arXiv:2412.18426},
  year={2024}
}

@article{hu2024auitestagent,
  title={Auitestagent: Automatic requirements oriented gui function testing},
  author={Hu, Yongxiang and Wang, Xuan and Wang, Yingchuan and Zhang, Yu and Guo, Shiyu and Chen, Chaoyi and Wang, Xin and Zhou, Yangfan},
  journal={arXiv preprint arXiv:2407.09018},
  year={2024}
}

@inproceedings{su2021benchmarking,
  title={Benchmarking automated gui testing for android against real-world bugs},
  author={Su, Ting and Wang, Jue and Su, Zhendong},
  booktitle={Proceedings of the 29th ACM Joint Meeting on European Software Engineering Conference and Symposium on the Foundations of Software Engineering},
  pages={119--130},
  year={2021}
}

@article{peng2023impact,
  title={The impact of ai on developer productivity: Evidence from github copilot},
  author={Peng, Sida and Kalliamvakou, Eirini and Cihon, Peter and Demirer, Mert},
  journal={arXiv preprint arXiv:2302.06590},
  year={2023}
}

@inproceedings{si2025design2code,
  title={Design2code: Benchmarking multimodal code generation for automated front-end engineering},
  author={Si, Chenglei and Zhang, Yanzhe and Li, Ryan and Yang, Zhengyuan and Liu, Ruibo and Yang, Diyi},
  booktitle={Proceedings of the 2025 Conference of the Nations of the Americas Chapter of the Association for Computational Linguistics: Human Language Technologies (Volume 1: Long Papers)},
  pages={3956--3974},
  year={2025}
}

@article{liu2025seeing,
  title={Seeing is believing: Vision-driven non-crash functional bug detection for mobile apps},
  author={Liu, Zhe and Li, Cheng and Chen, Chunyang and Wang, Junjie and Chen, Mengzhuo and Wu, Boyu and Wang, Yawen and Hu, Jun and Wang, Qing},
  journal={IEEE Transactions on Software Engineering},
  year={2025},
  publisher={IEEE}
}

@inproceedings{liu2024make,
  title={Make llm a testing expert: Bringing human-like interaction to mobile gui testing via functionality-aware decisions},
  author={Liu, Zhe and Chen, Chunyang and Wang, Junjie and Chen, Mengzhuo and Wu, Boyu and Che, Xing and Wang, Dandan and Wang, Qing},
  booktitle={Proceedings of the IEEE/ACM 46th International Conference on Software Engineering},
  pages={1--13},
  year={2024}
}

@article{yoon2023autonomous,
  title={Autonomous large language model agents enabling intent-driven mobile gui testing},
  author={Yoon, Juyeon and Feldt, Robert and Yoo, Shin},
  journal={arXiv preprint arXiv:2311.08649},
  year={2023}
}

@article{barr2014oracle,
  title={The oracle problem in software testing: A survey},
  author={Barr, Earl T and Harman, Mark and McMinn, Phil and Shahbaz, Muzammil and Yoo, Shin},
  journal={IEEE transactions on software engineering},
  volume={41},
  number={5},
  pages={507--525},
  year={2014},
  publisher={IEEE}
}

@inproceedings{zhou2024webarena,
  title={Webarena: A realistic web environment for building autonomous agents},
  author={Zhou, Shuyan and Xu, Frank F and Zhu, Hao and Zhou, Xuhui and Lo, Robert and Sridhar, Abishek and Cheng, Xianyi and Ou, Tianyue and Bisk, Yonatan and Fried, Daniel and others},
  booktitle={International Conference on Learning Representations},
  volume={2024},
  pages={15585--15606},
  year={2024}
}

@article{xie2024osworld,
  title={Osworld: Benchmarking multimodal agents for open-ended tasks in real computer environments},
  author={Xie, Tianbao and Zhang, Danyang and Chen, Jixuan and Li, Xiaochuan and Zhao, Siheng and Cao, Ruisheng and Hua, Toh J and Cheng, Zhoujun and Shin, Dongchan and Lei, Fangyu and others},
  journal={Advances in Neural Information Processing Systems},
  volume={37},
  pages={52040--52094},
  year={2024}
}

@inproceedings{rawles2025androidworld,
  title={Androidworld: A dynamic benchmarking environment for autonomous agents},
  author={Rawles, Chris and Clinckemaillie, Sarah and Chang, Yifan and Waltz, Jonathan and Lau, Gabrielle and Fair, Marybeth and Li, Alice and Bishop, William and Li, Wei and Campbell-Ajala, Folawiyo and others},
  booktitle={International Conference on Learning Representations},
  volume={2025},
  pages={406--441},
  year={2025}
}

@article{memon2001hierarchical,
  title={Hierarchical GUI test case generation using automated planning},
  author={Memon, Atif M and Pollack, Martha E and Soffa, Mary Lou},
  journal={IEEE transactions on software engineering},
  volume={27},
  number={2},
  pages={144--155},
  year={2001},
  publisher={IEEE}
}

@article{banerjee2013graphical,
  title={Graphical user interface (GUI) testing: Systematic mapping and repository},
  author={Banerjee, Ishan and Nguyen, Bao and Garousi, Vahid and Memon, Atif},
  journal={Information and Software Technology},
  volume={55},
  number={10},
  pages={1679--1694},
  year={2013},
  publisher={Elsevier}
}

@inproceedings{choudhary2015automated,
  title={Automated test input generation for android: Are we there yet?(e)},
  author={Choudhary, Shauvik Roy and Gorla, Alessandra and Orso, Alessandro},
  booktitle={2015 30th IEEE/ACM International Conference on Automated Software Engineering (ASE)},
  pages={429--440},
  year={2015},
  organization={IEEE}
}

@inproceedings{mao2016sapienz,
  title={Sapienz: Multi-objective automated testing for android applications},
  author={Mao, Ke and Harman, Mark and Jia, Yue},
  booktitle={Proceedings of the 25th international symposium on software testing and analysis},
  pages={94--105},
  year={2016}
}

@misc{monkey2020,
  author={{Monkey}},
  title={Monkey},
  year={2020},
  howpublished={\url{http://developer.android.com/tools/help/monkey.html}},
  note={Retrieved 2026-09}
}

@inproceedings{just2014mutants,
  title={Are mutants a valid substitute for real faults in software testing?},
  author={Just, Ren{\'e} and Jalali, Darioush and Inozemtseva, Laura and Ernst, Michael D and Holmes, Reid and Fraser, Gordon},
  booktitle={Proceedings of the 22nd ACM SIGSOFT International Symposium on Foundations of Software Engineering},
  pages={654--665},
  year={2014}
}

@article{chen2026guitestscape,
  title={GUITestScape: Towards Open-set Evaluation on Exploratory GUI Testing},
  author={Chen, Xiaoyi and Gao, Yifei and Xu, Yang and Song, Xingxing and Zhang, Yi and Sang, Jitao},
  journal={arXiv preprint arXiv:2605.29532},
  year={2026}
}

@article{wang2026cua,
  title={Cua-gym: Scaling verifiable training environments and tasks for computer-use agents},
  author={Wang, Bowen and Lu, Dunjie and Wang, Junli and Bai, Tianyi and Liu, Shixuan and Zhang, Zhipeng and Wang, Haiquan and Hu, Hao and Xie, Tianbao and Bai, Shuai and others},
  journal={arXiv preprint arXiv:2605.25624},
  year={2026}
}

@article{zhuge2024agent,
  title={Agent-as-a-judge: Evaluate agents with agents},
  author={Zhuge, Mingchen and Zhao, Changsheng and Ashley, Dylan and Wang, Wenyi and Khizbullin, Dmitrii and Xiong, Yunyang and Liu, Zechun and Chang, Ernie and Krishnamoorthi, Raghuraman and Tian, Yuandong and others},
  journal={arXiv preprint arXiv:2410.10934},
  year={2024}
}

@article{zheng2023judging,
  title={Judging llm-as-a-judge with mt-bench and chatbot arena},
  author={Zheng, Lianmin and Chiang, Wei-Lin and Sheng, Ying and Zhuang, Siyuan and Wu, Zhanghao and Zhuang, Yonghao and Lin, Zi and Li, Zhuohan and Li, Dacheng and Xing, Eric and others},
  journal={Advances in neural information processing systems},
  volume={36},
  pages={46595--46623},
  year={2023}
}

@inproceedings{just2014defects4j,
  title={Defects4J: A database of existing faults to enable controlled testing studies for Java programs},
  author={Just, Ren{\'e} and Jalali, Darioush and Ernst, Michael D},
  booktitle={Proceedings of the 2014 international symposium on software testing and analysis},
  pages={437--440},
  year={2014}
}

@inproceedings{patra2021semantic,
  title={Semantic bug seeding: a learning-based approach for creating realistic bugs},
  author={Patra, Jibesh and Pradel, Michael},
  booktitle={Proceedings of the 29th ACM Joint Meeting on European Software Engineering Conference and Symposium on the Foundations of Software Engineering},
  pages={906--918},
  year={2021}
}

@inproceedings{ibrahimzada2025challenging,
  title={Challenging bug prediction and repair models with synthetic bugs},
  author={Ibrahimzada, Ali Reza and Chen, Yang and Rong, Ryan and Jabbarvand, Reyhaneh},
  booktitle={2025 IEEE International Conference on Source Code Analysis \& Manipulation (SCAM)},
  pages={133--144},
  year={2025},
  organization={IEEE}
}

@inproceedings{hong2024cogagent,
  title={Cogagent: A visual language model for gui agents},
  author={Hong, Wenyi and Wang, Weihan and Lv, Qingsong and Xu, Jiazheng and Yu, Wenmeng and Ji, Junhui and Wang, Yan and Wang, Zihan and Dong, Yuxiao and Ding, Ming and others},
  booktitle={2024 IEEE/CVF Conference on Computer Vision and Pattern Recognition (CVPR)},
  pages={14281--14290},
  year={2024},
  organization={IEEE}
}

@article{qin2025ui,
  title={Ui-tars: Pioneering automated gui interaction with native agents},
  author={Qin, Yujia and Ye, Yining and Fang, Junjie and Wang, Haoming and Liang, Shihao and Tian, Shizuo and Zhang, Junda and Li, Jiahao and Li, Yunxin and Huang, Shijue and others},
  journal={arXiv preprint arXiv:2501.12326},
  year={2025}
}

@article{wang2026liveevalbench,
  title={{LiveEvalBench}: Toward open-world evaluation for web generation},
  author={Wang, Yiyao and Wen, Zhen and Tang, Yinghao and others},
  journal={arXiv preprint arXiv:2608.03689},
  year={2026}
}

@article{yang2026cookiebench,
  title={Cookie-Bench: Continuous on-screen key interaction evaluation for web generation},
  author={Yang, Haoyue and Shen, Zhangxiao and Ding, Fan and others},
  journal={arXiv preprint arXiv:2605.30000},
  year={2026}
}

@article{peng2026playcoder,
  title={{PlayCoder}: Making {LLM}-generated {GUI} code playable},
  author={Peng, Zhiyuan and Tao, Wei and Yin, Xin and Ying, Chenhao and Luo, Yuan and Guo, Yiwen},
  journal={arXiv preprint arXiv:2604.19742},
  year={2026}
}

@book{zeldman2001talent,
  title     = {Taking Your Talent to the Web: A Guide for the Transitioning Designer},
  author    = {Zeldman, Jeffrey},
  year      = {2001},
  publisher = {New Riders Publishing},
  address   = {Indianapolis, IN},
}

@inproceedings{yao2023react,
  title = {ReAct: Synergizing Reasoning and Acting in Language Models},
  author = {Yao, Shunyu and Zhao, Jeffrey and Yu, Dian and Du, Nan and Shafran, Izhak and Narasimhan, Karthik and Cao, Yuan},
  booktitle = {International Conference on Learning Representations},
  year = {2023},
  url = {https://arxiv.org/abs/2210.03629}
}

@misc{anthropic2026claudecode,
  title = {Claude Code Documentation: Model Configuration},
  author = {{Anthropic}},
  year = {2026},
  url = {https://code.claude.com/docs/en/model-config},
  note = {Accessed: 2026-10-30}
}

@misc{anthropic2026opus55,
  title = {Claude Opus 5.5 System Card},
  author = {{Anthropic}},
  year = {2026},
  url = {https://www.anthropic.com/system-cards},
  note = {Accessed: 2026-10-30}
}

@article{xu2026deepseek,
  title={Deepseek-v4: Towards highly efficient million-token context intelligence},
  author={Xu, Anyi and Lin, Bangcai and Xue, Bing and Wang, Bingxuan and Xu, Bingzheng and Wu, Bochao and Zhang, Bowei and Lin, Chaofan and Dong, Chen and Ling, Chenchen and others},
  journal={arXiv preprint arXiv:2606.19348},
  year={2026}
}

@misc{google2026gemini38flash,
  title = {Gemini 3.8 Flash Model Card}, author = {{Google DeepMind}}, year = {2026},
  url = {https://deepmind.google/models/model-cards/gemini-3-8-flash/}}

@misc{qwen2026qwen37plus,
  title = {Qwen3.7-Plus}, author = {{Qwen Team}}, year = {2026},
  url = {https://qwen.ai/blog?id=qwen3.7-plus}}

@misc{bytedance2026seed21,
  title = {Seed2.1 Officially Released}, author = {{ByteDance Seed}}, year = {2026},
  url = {https://seed.bytedance.com/en/blog/seed2-1-officially-released-advancing-ai-productivity}}

@misc{zai2026glm53flash,
  title = {GLM-5.3-Flash}, author = {{Z.ai}}, year = {2026},
  url = {https://z.ai/blog/glm-5.3-flash}}

@inproceedings{hong2026framework,
  title={Framework and Benchmark for Code-Driven Agentic Testing in Web Development},
  author={Hong, Bin and Zhang, Zhenchao and He, Jiyuan and Zhang, Kai and Huang, Zhenya},
  booktitle={Proceedings of the 2026 Conference on Empirical Methods in Natural Language Processing},
  pages={47250--47279},
  year={2026}
}

@article{teoh2026webtestpilot,
  title={Webtestpilot: Agentic end-to-end web testing against natural language specification by inferring oracles with symbolized gui elements},
  author={Teoh, Xiwen and Lin, Yun and Nguyen, Duc-Minh and Ren, Ruofei and Zhang, Wenjie and Dong, Jin Song},
  journal={Proceedings of the ACM on Software Engineering},
  volume={3},
  number={FSE},
  pages={1933--1956},
  year={2026},
  publisher={ACM, New York, NY, USA}
}
\clearpage
\appendix
\setlength{\textfloatsep}{\supptextfloatsep}
\setlength{\floatsep}{\suppfloatsep}
\setlength{\intextsep}{\suppintextsep}
\setlength{\abovecaptionskip}{\suppabovecaptionskip}
\setlength{\belowcaptionskip}{\suppbelowcaptionskip}
\setcounter{table}{0}
\setcounter{figure}{0}
\renewcommand{\thetable}{A\arabic{table}}
\renewcommand{\thefigure}{A\arabic{figure}}
\twocolumn[{\centering\Large\bfseries Appendix\par\vspace{1em}}]
\section{Defect Synthesis Pipeline}
\label{sec:supp_pipeline}

This section expands the two-stage defect synthesis pipeline described in Section~3 of the main paper.
The first stage distills a corpus of real bug fixes into a library of reusable defect
patterns. The second stage turns a sampled pattern into a defective instance. It
identifies injectable sites in the source code, asks an LLM to write a contextual code edit and a defect
account, verifies the defect by differential replay on a clean and a defective instance, and
samples the verified defects to match the target distribution. Figure~2 of the main
paper summarizes this flow, with pattern distillation in (a), site identification in
(b), injection in (c), and verification in (d). The rest of this section gives the
implementation details behind each step.

\subsection{Stage 1: Distilling Defect Patterns}
\label{sec:supp_distill}

\begin{figure}[t]
  \centering
  \includegraphics[width=\columnwidth]{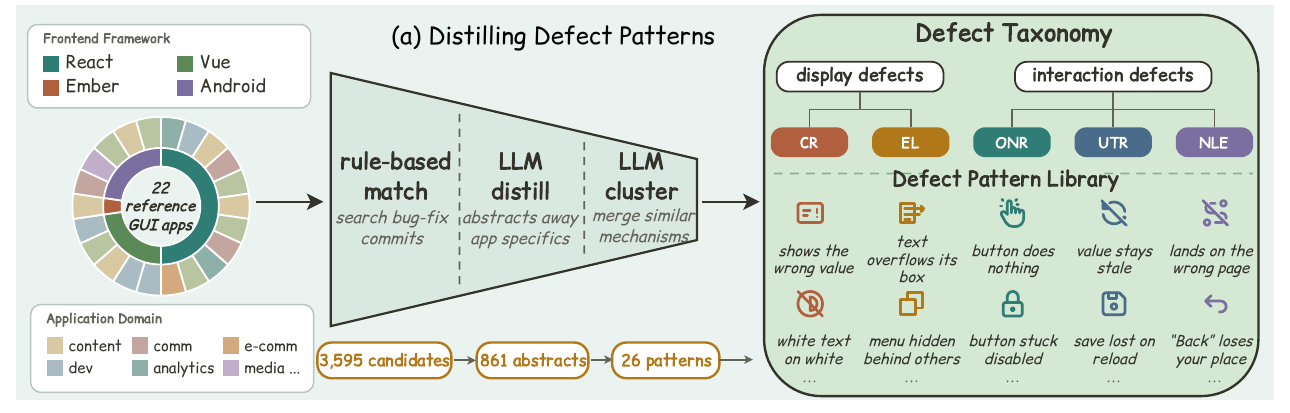}
  \caption{Pattern distillation, reproduced from panel (a) of Figure~2 in the main
  paper. We mine real bug-fix commits and distill them into a library of defect
  patterns.}
  \label{fig:supp_fig2a}
\end{figure}

Figure~\ref{fig:supp_fig2a} reproduces panel (a) of Figure~2 in the main paper and
summarizes this stage. We first build the corpus of real bug-fix commits, then
abstract them into defect patterns for reusable defect injection at the next stage.

\tightpara{Corpus}
We distill patterns from 22 public GitHub repositories. The corpus was curated to
satisfy four requirements. Each repository has a GUI-heavy frontend in React, Vue,
Ember, or Android, so the mined fixes exercise the kind of interface the benchmark
targets. Each has a substantial history of bug-fix commits, including GUI-level
regression tests for the interaction channel and display-related fix messages for the
display channel. The set covers the application domains used by prior GUI-agent
benchmarks, namely analytics, project management, content publishing, developer
tools, commerce, communication, and media/mobile. Finally, the repositories differ
enough in size and domain that no single project dominates the pattern library.
Table~\ref{tab:supp_repos} lists the repositories, their domains, their frontend
frameworks, their GitHub stars, and their lines of code (LOC).

\begin{table*}[t]
  \centering
  \small
  \setlength{\tabcolsep}{4pt}
  \caption{The 22 open-source repositories used for pattern distillation. Stars were
  retrieved from the GitHub API at the time of writing. LOC is counted on the
  repository snapshot used for mining.}
  \label{tab:supp_repos}
  \begin{tabular}{p{3.0cm}p{2.8cm}lrr}
    \toprule
    Repository & Domain & Framework & Stars (k) & LOC (k) \\
    \midrule
    metabase          & Analytics      & React   & 50.4  & 2{,}134 \\
    grafana           & Analytics      & React   & 78.0  & 4{,}323 \\
    Ghost             & Content        & Ember   & 55.8  & 1{,}178 \\
    gutenberg         & Content        & React   & 11.8  & 1{,}226 \\
    directus          & Content        & Vue     & 38.7  & 599 \\
    twenty            & Developer tools & React   & 61.1  & 2{,}801 \\
    cal.com           & Developer tools & React   & 50.1  & 698 \\
    vikunja           & Developer tools & Vue     & 6.1   & 429 \\
    AFFiNE            & Developer tools & React   & 74.6  & 1{,}098 \\
    appsmith          & Developer tools & React   & 41.3  & 1{,}833 \\
    n8n               & Developer tools & Vue     & 210.9 & 5{,}325 \\
    gitea             & Developer tools & Vue     & 59.1  & 616 \\
    medusa            & E-commerce     & React   & 37.3  & 6{,}385 \\
    nocodb            & E-commerce     & Vue     & 65.6  & 2{,}707 \\
    mattermost        & Communication  & React   & 39.6  & 2{,}684 \\
    Rocket.Chat       & Communication  & React   & 46.6  & 1{,}144 \\
    Signal-Android    & Communication  & Android & 29.7  & 1{,}258 \\
    excalidraw        & Media          & React   & 135.9 & 248 \\
    AntennaPod        & Media          & Android & 8.3   & 119 \\
    Anki-Android      & Media          & Android & 12.3  & 368 \\
    Wikipedia-Android & Media          & Android & 3.1   & 396 \\
    OpenTracks        & Media          & Android & 1.4   & 60 \\
    \bottomrule
  \end{tabular}
\end{table*}

\tightpara{Mining}
We use two channels because interaction and display bugs leave different traces. The
interaction channel scans commits whose subject starts with
\texttt{fix} and that touch a test file. A commit is kept when the added test code
contains at least three GUI actions and one
assertion. The regression test is an already-minimized, machine-readable
reproduction whose actions map onto the traversal action space. The channel yields
1,893 candidates.

The display channel targets the bugs that the interaction channel misses. Display
bugs are visible on arrival and rarely ship a GUI test, so we scan \texttt{fix}
commits whose subject signals a layout, rendering, or styling problem, without
requiring a test file, and use the added lines of the non-test source diff as the
signal. This channel yields 1,750
candidates. The two sets are de-duplicated by repository and commit. When a commit
is matched by both channels, the interaction form is kept because its regression
test is the stronger signal. The corpus contains 3,595 candidates.

\tightpara{Abstraction}
We sample at most 45 candidates per repository, stratified by the length of
their triggering sequence, and ask a text-only LLM (Sonnet 5.5) to turn each
candidate into an application-agnostic abstraction. An abstraction records three things: the
expected behavior it breaks, the interaction that exposes it, and a \emph{code
signature}, the structural code pattern that locates injectable sites in a new
application. The prompt forbids application-specific nouns in the first two
fields and requires a code signature specific enough to search source code;
candidates whose root cause lies outside the frontend are dropped. This leaves
861 frontend abstractions, one per mined fix.

A second pass (Sonnet 5.5) groups these abstractions into the 26 patterns used for
efficient injection, under the five fixed categories. The vocabulary of the first pass was
written for interaction bugs, and the display channel introduced candidates that
it does not describe, so the grouping is induced from the abstractions
themselves rather than from a fixed list. For each pattern the catalog records its category, the
number of real fixes behind it, its source repositories, its code signatures,
the length range of its triggering sequence, and the share of its mined
instances whose symptom appeared away from the trigger screen, i.e., its
near/far split. The injection stage reads only this catalog: the code
signature decides which sites can host a pattern, and the split sets the
pattern's target symptom distance. Table~\ref{tab:supp_pattern_example} shows
one example pattern from each channel.

\begin{table*}[t]
  \centering
  \small
  \setlength{\tabcolsep}{4pt}
  \caption{\textbf{Two example patterns.}
  \emph{Source} indicates the origin of each pattern, including the number of bug fixes and repositories from which it was derived.
  \emph{Correct behavior} describes what the app should do before the defect is introduced.
  \emph{Triggering interaction} describes the user action that can reveal the defect and what to check afterward.
  \emph{Code signature} describes characteristic code structures associated with the pattern.}
  \label{tab:supp_pattern_example}
  \begin{tabular}{>{\raggedright\arraybackslash}p{2.6cm}
                  >{\raggedright\arraybackslash}p{6.4cm}
                  >{\raggedright\arraybackslash}p{6.4cm}}
    \toprule
    & \textbf{stacking\_and\_occlusion} (EL) & \textbf{detached\_or\_missing\_handler} (ONR) \\
    \midrule
    Source
      & 34 fixes from 18 repositories
      & 15 fixes from 12 repositories \\[4pt]
    Correct behavior
      & Opening a menu shows all its options.
      & Clicking a button or link performs the expected action. \\[4pt]
    Triggering interaction
      & Open a menu and check whether any options are cut off or covered.
      & Click a button or link and check whether it performs the expected action,
        such as opening a window or saving the entered information. \\[4pt]
    Code signature
      & Menu components nested inside a container with CSS
        \texttt{overflow: hidden} or \texttt{overflow: auto}, or assigned
        a lower \texttt{z-index} than overlapping elements.
      & Button components with a missing \texttt{onClick} or
        \texttt{onSubmit} binding, or an event handler that fails to
        call the supplied callback. \\
    \bottomrule
  \end{tabular}
\end{table*}

\subsection{Stage 2: From Patterns to Defective Instances}
\label{sec:supp_stage2}

The second stage identifies the sites where a pattern can be injected, writes the
edit and the defect account, verifies the account by execution, and composes the
admitted defects into one instance per application. The first three steps
correspond to Figure~2(b)--(d) of the main paper; sampling follows Section~4, and
composition is described at the end of this subsection.

\subsubsection{Identifying injectable sites}
\label{sec:supp_sites}

We build an instrumented copy of each application whose only difference from the
original is a stable identifier stamped on every event handler and dynamic text
binding that can render a UI element. The identifier surfaces in the DOM at
runtime, so a site in the source can be matched to the element it renders, and
the copy serves as the clean instance for injection and replay.

A crawler then explores the app from its entry page and records an interaction
graph whose nodes are UI states and whose edges are clicks, the action the
verifier can replay. The crawl is a rule-based search: a Playwright driver tries each actionable element at most once,
expands the frontier breadth-first, and replays each path from a fresh server
reset and empty browser storage, so every recorded path can be executed again.
The length of a site's recorded path is its \emph{navigation depth}. This
instrumentation process also enables to evaluate testing-agent coverage in
our experiments.

\subsubsection{Injecting defects}
\label{sec:supp_inject}

A candidate injection is a pair of an injectable site and a pattern, and it is
feasible when the site's source file contains one of the pattern's code
signatures. Pairs are sampled to match the target distribution
(Sec.~\ref{sec:supp_sampling}).

For each pair the LLM (Sonnet 5.5) receives the pattern with its code signatures and a few
real fixes, together with the source file and the reachable sites in it, grouped
by page and annotated with their depth. It returns a minimal contextual edit for defect injection and the corresponding defect account containing:
\begin{itemize}
\item \texttt{correctBehavior}: the correct behavior before injection, stated
  without naming the defect;
\item \texttt{symptom}: the symptom a user would observe after it, and
  \texttt{symptomDistance}, which records whether the symptom appears on the
  screen where the action is triggered or only after leaving it;
\item \texttt{reproductionScript}: a short sequence of clicks, typing, and
  reloads on named sites, with navigation taken from the recorded action paths.
\end{itemize}
The defect category is inherited from the corresponding pattern category. If the file cannot accommodate the pattern, the LLM says so rather than force an unconvincing edit. At this stage, the account remains a set of claims, which are verified in the next step.

\subsubsection{Verifying defects by execution}
\label{sec:supp_verify}

Verification boots a clean and a defective instance from the same seed state and
replays the reproduction script on both, comparing the rendered page element by
element. If the scripted actions leave the two instances identical, the symptom
may still appear later or on another screen. The verifier therefore looks further
before giving up: it opens the screen declared for the symptom, navigates away and
returns, and reloads the page. Navigating away and back exposes state that was
never committed to the app's store, and reloading exposes changes that appeared
saved but were never persisted. Each failure is recorded with its cause, so it
can be repaired in the right place, and a defect hidden by another in a composed
instance is not counted as a miss.

Scripted replay breaks whenever a recorded selector stops matching the page, which
the injected edit itself can cause. An agent (Sonnet 5.5) then repairs only the
failed step and stops where the symptom should be visible; a vision-language model
(Gemini 3.1 Pro) compares the clean and defective screenshots and admits the
defect only if it can name the difference that matches the declared symptom. A
failed check then returns its diagnosis to the LLM, which re-grounds the edit, for
up to three measured attempts.

\subsubsection{Sampling and composing the released set}
\label{sec:supp_sampling}

\tightpara{Sampling}
We inject in rounds so that the released defects match the target distribution over
category, navigation depth, and symptom distance (Sec.~4 of the main paper). Each
application plans 12 defects, a site is used at most once, and a pattern at most
five times per application. Verification discards defects unevenly, so after each
round the planner counts the manifested defects per cell and gives the next round's
slots only to categories that are still short; a cell that stays short is recorded as a
shortfall rather than filled off-target.

\tightpara{Composition}
A traversal episode gives the agent one application, so the released unit is one
composed instance per application. Selection from the verified pool uses the same
three axes, prefers the application and the page with the fewest selected defects,
and is deterministic given the same records, so defects spread across applications
and pages. Composition applies the selected defects in a fixed order and re-checks
uniqueness and parseability after each substitution. Single-defect manifestation
does not carry over to a composed instance, since one defect can lock the navigation
or overwrite the state another needs, so the pipeline re-runs every defect's own
reproduction script on the composed instance and drops masked or blocked defects,
repeating until every remaining defect manifests. Blocked, dud, or crashing
defects are removed from the final roster with a recorded reason.

\section{Benchmark Details}
\label{sec:supp_benchmark}

The released benchmark contains 20 web applications, 10 real open-source
projects and 10 mock apps of commercial products~\citep{wang2026cua}, whose sources range from
5.8k to 41.8k lines (Table~\ref{tab:supp_apps}). Every app carries between two
and eleven injected defects, 95 in total: 31 unexpected task results, 22
element-layout, 19 operation-no-response, 17 content-rendering, and 6
navigation-logic-error defects. The two groups differ in scale and in defect
mix: the real projects are roughly three times larger in code (163k vs.\ 55k
lines) yet carry fewer defects (40 vs.\ 55), while the mocks concentrate
interaction defects (38 vs.\ 18) and deep defects (21 vs.\ 2) in much smaller
codebases.

Perhaps surprisingly, the mock applications provide more injectable sites despite having fewer lines of code than the reals.
This difference may reflect how the
AI-generated mocks reproduce commercial interfaces: they implement many
visible controls and interactions with relatively little supporting logic.
Real projects may devote more code to reusable components, utilities, and
internal logic that do not directly add injectable sites.

\begin{table*}[t]
  \centering
  \small
  \setlength{\tabcolsep}{4pt}
  \caption{\textbf{Benchmark applications.} \emph{Real} apps are open-source
  projects; \emph{mock} apps reproduce the frontend of commercial products.
  \emph{\#Sites} counts injectable sites after excluding those that are structurally
  unobservable, and is the denominator of Coverage.
  \emph{\#Defect} is the number of injected defects, split by category (CR: content
  rendering; EL: element layout; ONR: operation no response; UTR: unexpected
  task result; NLE: navigation logic error), by navigation depth (\emph{Deep}:
  three or more actions from the entry page), and by symptom distance
  (\emph{Far}: visible only away from the trigger screen).}
  \label{tab:supp_apps}
  \begin{tabular}{l l r r | r r r r r r r r}
    \toprule
    App & Domain & LOC (k) & \#Sites & \#Defect & CR & EL & ONR & UTR & NLE
        & Deep & Far \\
    \midrule
    \multicolumn{12}{l}{\emph{Real open-source projects (10)}} \\
    \texttt{drawdb}            & Developer tools & 41.8 & 249 & 3 & 2 & 0 & 1 & 0 & 0 & 0 & 0 \\
    \texttt{it\_tools}         & Developer tools & 16.9 & 352 & 5 & 3 & 1 & 0 & 1 & 0 & 0 & 0 \\
    \texttt{coreui\_vue}       & Admin dashboard & 13.9 &  51 & 4 & 0 & 2 & 1 & 1 & 0 & 0 & 2 \\
    \texttt{materio\_admin}    & Admin dashboard & 20.2 & 132 & 2 & 2 & 0 & 0 & 0 & 0 & 0 & 0 \\
    \texttt{shadcn\_admin}     & Admin dashboard & 17.8 &  96 & 8 & 0 & 2 & 2 & 1 & 3 & 2 & 1 \\
    \texttt{soybean\_admin}    & Admin dashboard & 10.1 &  92 & 3 & 1 & 0 & 0 & 2 & 0 & 0 & 0 \\
    \texttt{tailadmin\_react}  & Admin dashboard &  7.9 &  56 & 4 & 2 & 2 & 0 & 0 & 0 & 0 & 0 \\
    \texttt{tailadmin\_vue}    & Admin dashboard &  7.5 &  97 & 6 & 0 & 3 & 1 & 2 & 0 & 0 & 1 \\
    \texttt{vue\_naive\_admin} & Admin dashboard &  5.8 & 123 & 3 & 0 & 1 & 0 & 2 & 0 & 0 & 1 \\
    \texttt{mindmap}           & Productivity    & 21.2 & 429 & 2 & 1 & 0 & 1 & 0 & 0 & 0 & 0 \\
    \midrule
    \multicolumn{2}{l}{\emph{Subtotal (real)}}
      & 163.1 & 1{,}677 & 40 & 11 & 11 & 6 & 9 & 3 & 2 & 5 \\
    \midrule
    \multicolumn{12}{l}{\emph{Mock applications of commercial products (10)}} \\
    \texttt{gmail\_mock}        & Communication   &  4.1 & 153 &  5 & 0 & 1 & 2 & 1 & 1 & 0 & 0 \\
    \texttt{aliyun\_mock}       & Cloud console   &  4.9 & 346 & 11 & 0 & 2 & 2 & 6 & 1 & 3 & 2 \\
    \texttt{cloudflare\_mock}   & Cloud console   &  6.7 & 320 &  5 & 1 & 0 & 2 & 2 & 0 & 3 & 0 \\
    \texttt{github\_mock}       & Developer tools &  7.5 & 312 &  7 & 2 & 1 & 0 & 4 & 0 & 6 & 1 \\
    \texttt{booking\_com\_mock} & E-commerce      &  4.7 & 120 &  4 & 1 & 1 & 2 & 0 & 0 & 0 & 0 \\
    \texttt{Zendesk\_mock}      & CRM \& support  &  5.6 & 243 &  5 & 0 & 1 & 1 & 3 & 0 & 2 & 0 \\
    \texttt{adp\_mock}          & HR \& payroll   &  5.3 & 339 &  6 & 0 & 3 & 1 & 2 & 0 & 2 & 0 \\
    \texttt{hubspot\_mock}      & CRM \& support  &  4.7 & 311 &  4 & 0 & 0 & 1 & 2 & 1 & 1 & 0 \\
    \texttt{docusign\_mock}     & Productivity    &  5.4 & 221 &  4 & 1 & 1 & 1 & 1 & 0 & 2 & 0 \\
    \texttt{notion\_mock}       & Productivity    &  6.2 & 219 &  4 & 1 & 1 & 1 & 1 & 0 & 2 & 0 \\
    \midrule
    \multicolumn{2}{l}{\emph{Subtotal (mock)}}
      & 55.0 & 2{,}584 & 55 & 6 & 11 & 13 & 22 & 3 & 21 & 3 \\
    \midrule
    \multicolumn{2}{l}{\textbf{Total}}
      & \textbf{218.1} & \textbf{4{,}261} & \textbf{95}
      & \textbf{17} & \textbf{22} & \textbf{19} & \textbf{31} & \textbf{6}
      & \textbf{23} & \textbf{8} \\
    \bottomrule
  \end{tabular}
\end{table*}

\subsection{Example Defects}
\label{sec:supp_examples}

\begin{figure*}[t]
  \centering
  \small
  \raisebox{-.5\height}{\includegraphics[width=0.44\textwidth]{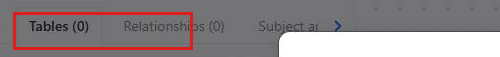}}
  \hspace{6pt}{\footnotesize\sffamily vs.}\hspace{6pt}
  \raisebox{-.5\height}{\includegraphics[width=0.44\textwidth]{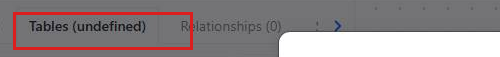}}\\[1pt]
  {\footnotesize \textbf{CR}, \texttt{drawdb}: the \textbf{Tables} tab shows \emph{0} in the clean app (left) and \emph{undefined} in the defective app (right).}\\[8pt]
  \raisebox{-.5\height}{\includegraphics[width=0.29\textwidth]{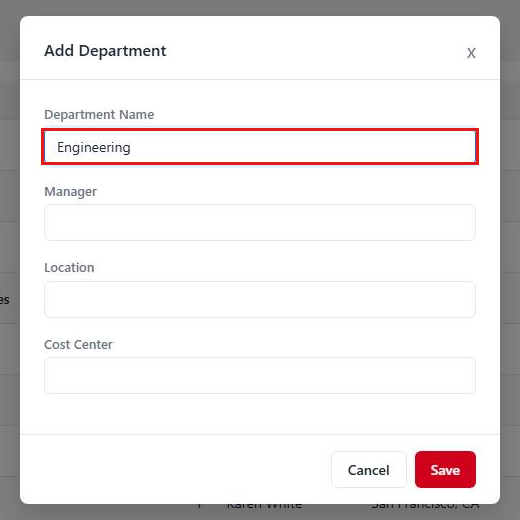}}
  \hspace{6pt}{\footnotesize\sffamily vs.}\hspace{6pt}
  \raisebox{-.5\height}{\includegraphics[width=0.29\textwidth]{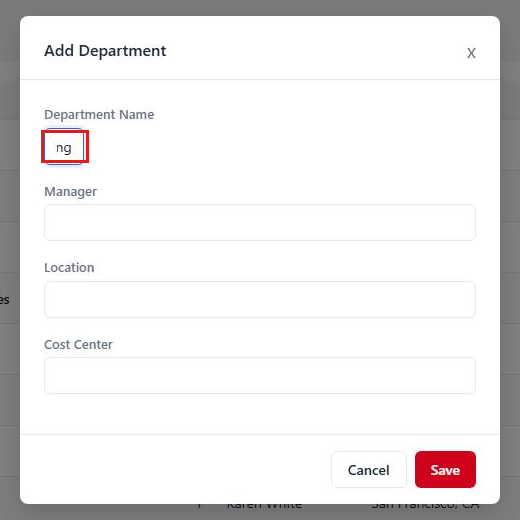}}\\[1pt]
  {\footnotesize \textbf{EL}, \texttt{adp\_mock}: the \textbf{Department Name} field has normal width in the clean app (left) but is too narrow to show its text in the defective app (right).}
  \caption{\textbf{Display defects.} Clean apps are on the left; defective apps are on the right. \textcolor[RGB]{220,30,30}{Red} boxes highlight the differences.}
  \label{fig:supp_display}
\end{figure*}

\textbf{Display defects} can be identified by checking the content and
layout of visible elements. \Cref{fig:supp_display} shows an incorrect
table count (CR) and a text field that is too narrow to display its
contents (EL)

\begin{figure*}[t]
  \centering
  \small
  \raisebox{-.5\height}{\includegraphics[width=0.38\textwidth]{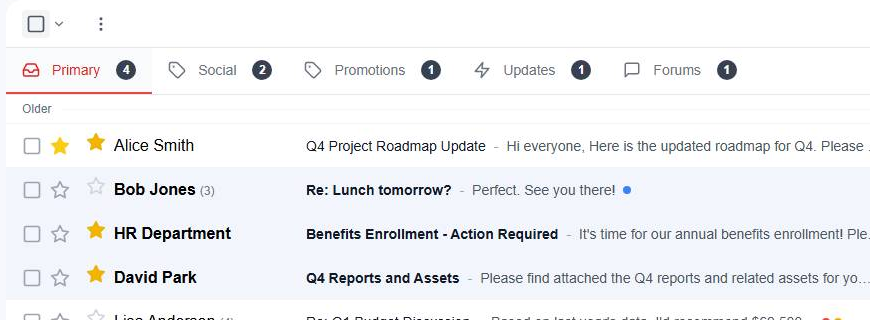}}
  \hspace{6pt}{\footnotesize\sffamily vs.}\hspace{6pt}
  \raisebox{-.5\height}{\includegraphics[width=0.38\textwidth]{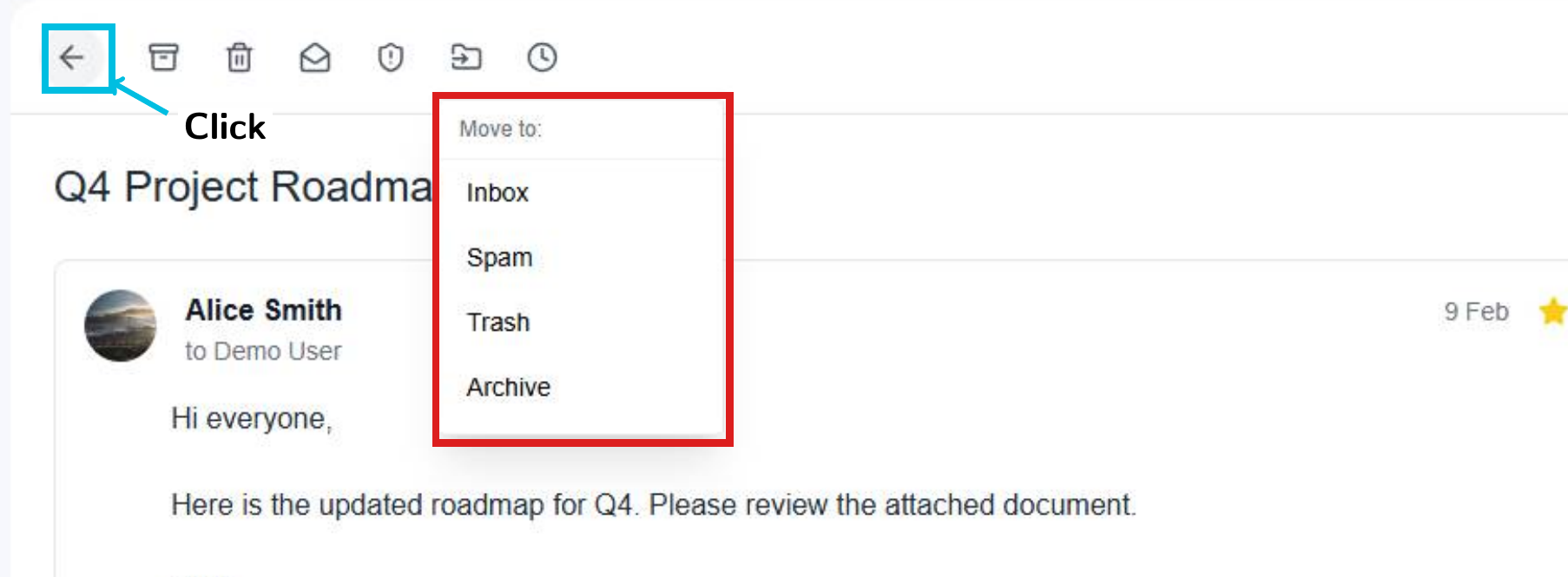}}\\[1pt]
  {\footnotesize \textbf{ONR}, \texttt{gmail\_mock}: Clicking \textbf{Back} icon returns to the inbox page in the clean app (left) but does nothing in the defective app (right).}\\[7pt]
  \raisebox{-.5\height}{\includegraphics[width=0.40\textwidth]{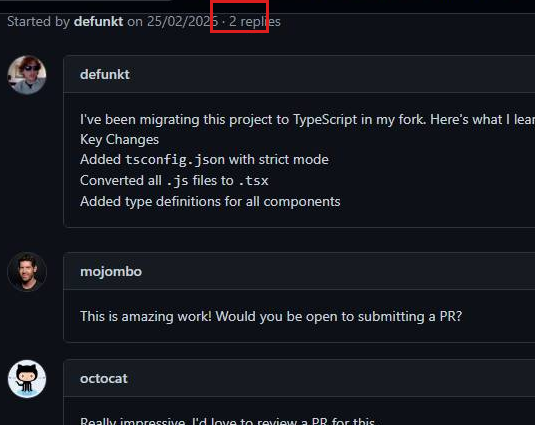}}
  \hspace{6pt}{\footnotesize\sffamily vs.}\hspace{6pt}
  \raisebox{-.5\height}{\includegraphics[width=0.40\textwidth]{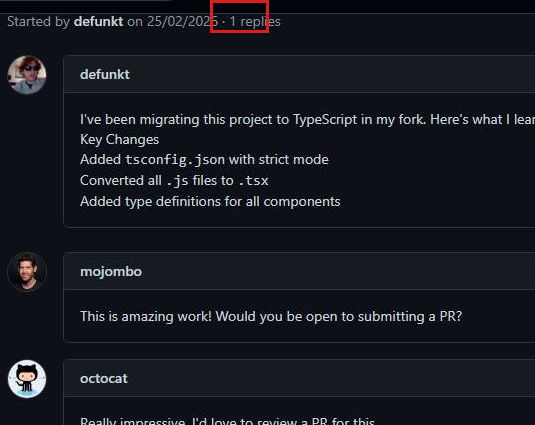}}\\[1pt]
  {\footnotesize \textbf{UTR}, \texttt{github\_mock}: after posting a new reply, the clean app counts \emph{2 replies} (left), while the defective app counts only \emph{1 reply} (right).}\\[7pt]
  \raisebox{-.5\height}{\includegraphics[width=0.42\textwidth]{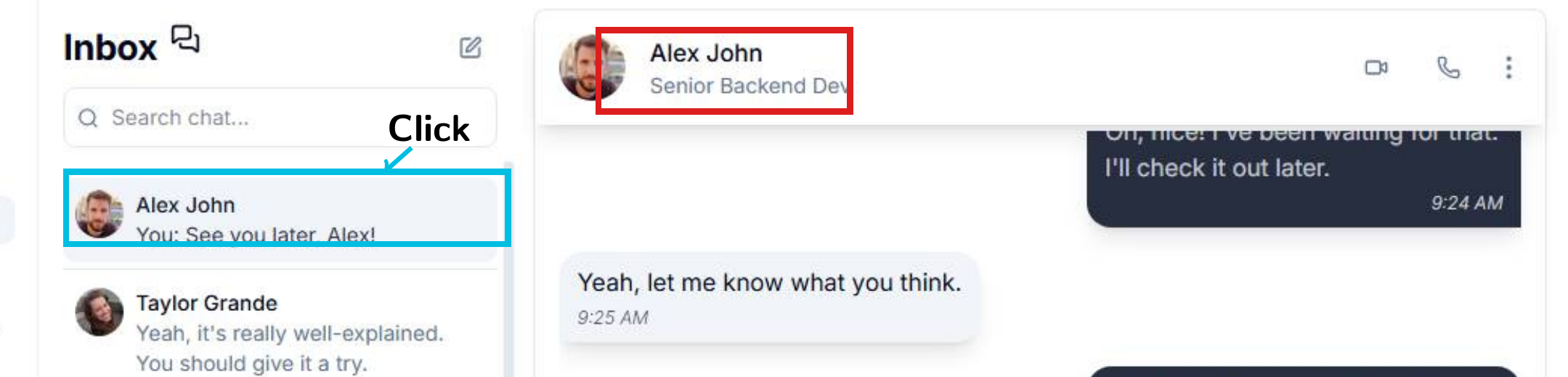}}
  \hspace{6pt}{\footnotesize\sffamily vs.}\hspace{6pt}
  \raisebox{-.5\height}{\includegraphics[width=0.42\textwidth]{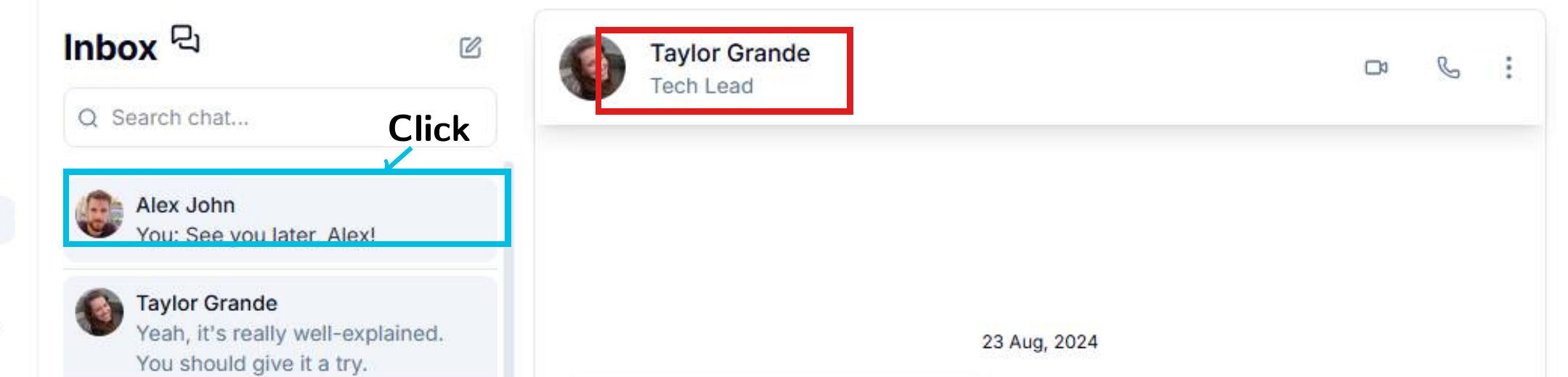}}\\[1pt]
  {\footnotesize \textbf{NLE}, \texttt{shadcn\_admin}: clicking \textbf{Alex John} opens his conversation in the clean app (left) but opens \textbf{Taylor Grande} in the defective app (right).}
  \caption{\textbf{Interaction defects.} Results of the same action in clean (left) and defective (right) apps. \textcolor[RGB]{0,190,225}{Cyan} boxes and \emph{Click} annotations mark click targets; \textcolor[RGB]{220,30,30}{red} boxes highlight the results being compared.}
  \label{fig:supp_interaction}
\end{figure*}

\textbf{Interaction defects} require checking whether an action produces
the expected result. \Cref{fig:supp_interaction} compares the results of
the same actions in clean and defective apps: clicking \emph{Back icon} has no
effect (ONR), posting a reply leaves the reply count incorrect (UTR),
and selecting a conversation opens a wrong one (NLE).

\begin{figure*}[!t]
  \centering
  \begin{minipage}[c]{0.34\textwidth}
    \centering
    \includegraphics[width=\linewidth]{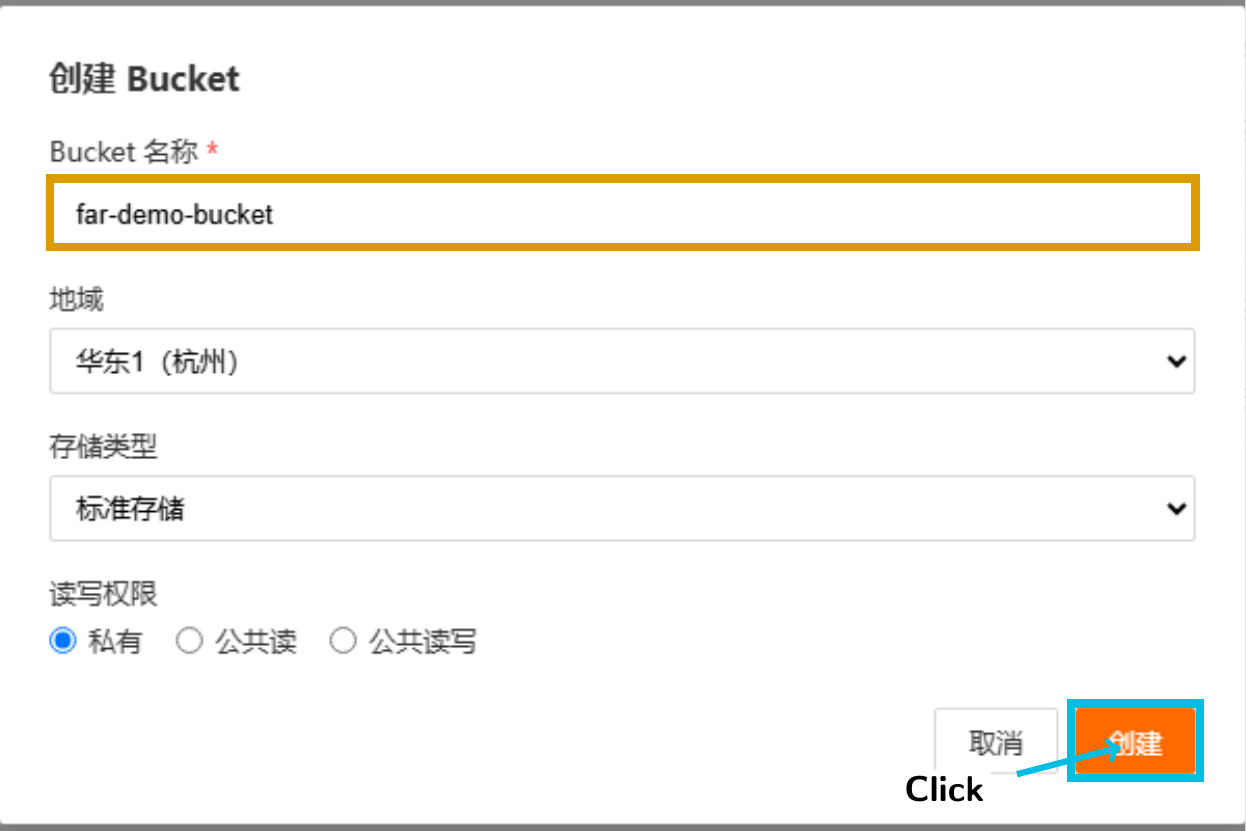}\\[2pt]
    {\footnotesize (a) Enter \emph{far-demo-bucket} as the bucket name and click \emph{Create}.}
  \end{minipage}\hfill
  $\longrightarrow$\hfill
  \begin{minipage}[c]{0.58\textwidth}
    \centering
    \includegraphics[width=\linewidth]{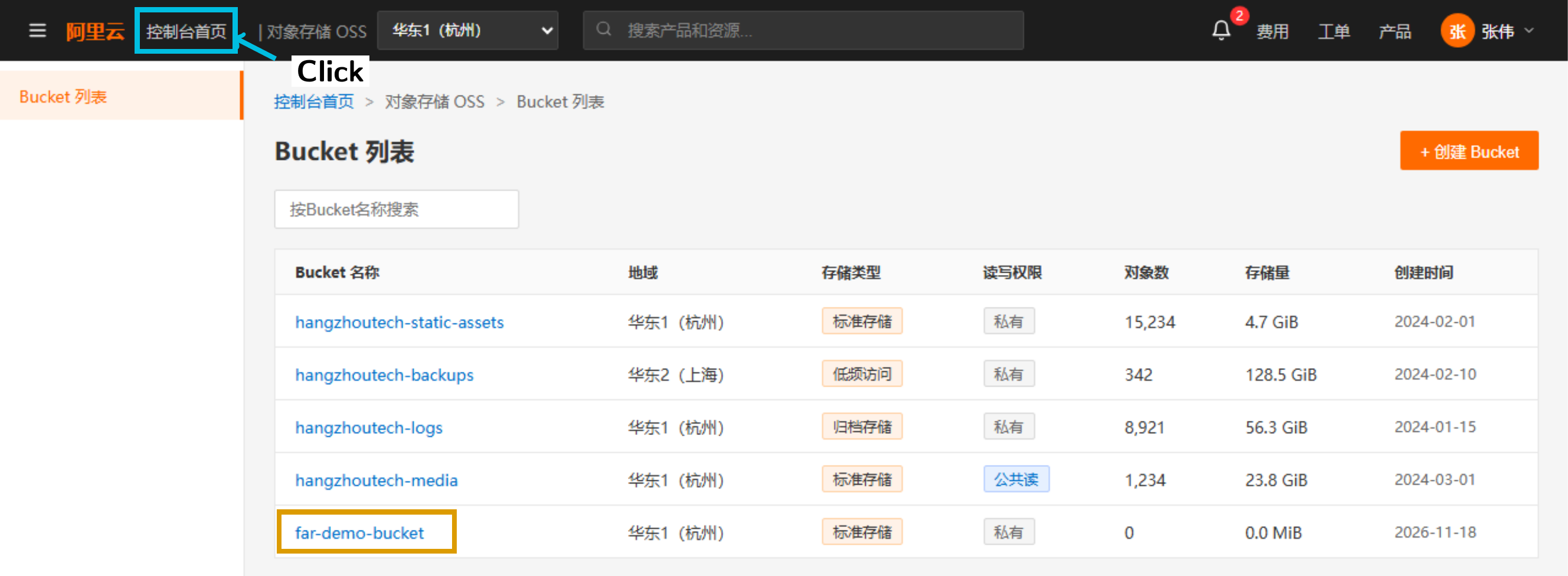}\\[2pt]
    {\footnotesize (b) The newly created bucket appears correctly in the bucket list. Then click \emph{Console Home}.}
  \end{minipage}\\[6pt]
  {\footnotesize Click Console Home $\downarrow$}\\[4pt]
  \includegraphics[width=0.98\textwidth]{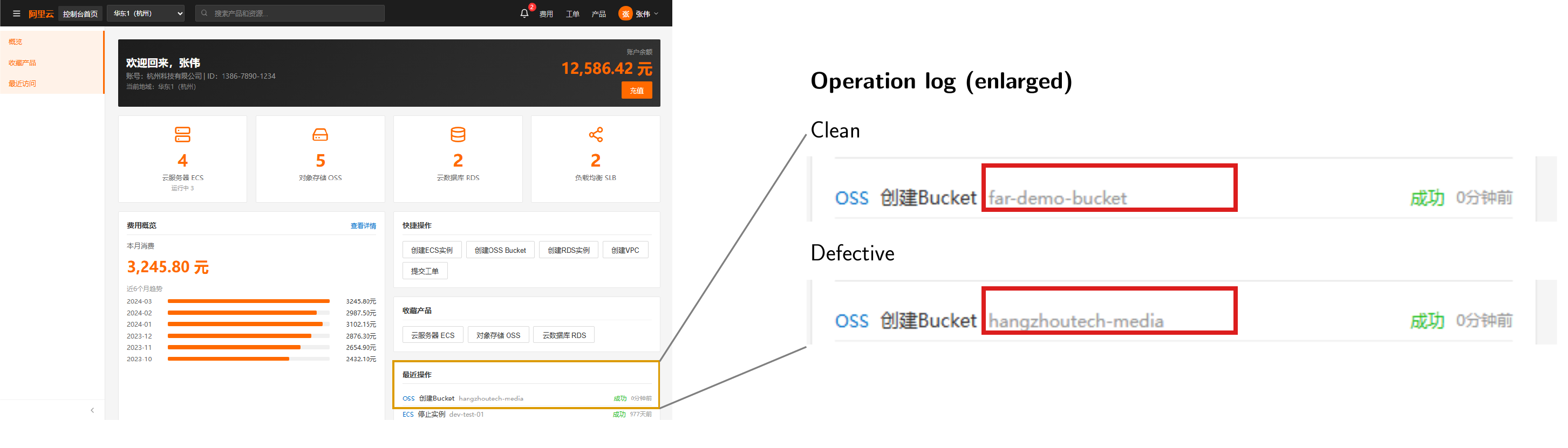}\\[2pt]
  {\footnotesize (c) On the \emph{Console Home} dashboard, compare the operation-log entries.}
  \caption{\textbf{A far defect} (\texttt{aliyun\_mock}).
  (a) The user creates a bucket named \emph{far-demo-bucket} using the form.
  (b) The bucket list shows the new entry correctly; the user then navigates to \emph{Console Home}.
  (c) Comparing the dashboard logs reveals the error: the clean app records \emph{far-demo-bucket},
  whereas the defective app records \emph{hangzhoutech-media}, which is not the newest record.
  The connecting lines link the log area to the enlarged entries.
  The symptom becomes visible only after navigating away from the creation screen.
  \textcolor[RGB]{0,190,225}{Cyan} boxes mark click targets;
  \textcolor[RGB]{220,155,0}{amber} boxes highlight information or regions to inspect;
  \textcolor[RGB]{220,30,30}{red} boxes highlight the values being compared between the clean and defective apps.}

  \label{fig:supp_far}
\end{figure*}

\textbf{Far defects} further require checking a different screen from the one
where the defect is triggered. In \Cref{fig:supp_far}, a newly created
bucket appears correctly in the list in both apps. After navigating
to the dashboard, however, the defective app's operation log shows
the wrong bucket name. Checking only the creation screen would miss
this defect.

\section{Additional Results}
\label{sec:supp_results}

\subsection{Setup Details}
\label{sec:supp_setup}

All episodes use the composed instances described in Section~\ref{sec:supp_benchmark} and the same set of 95 injected defects. Both harnesses receive only screenshots and output pixel-coordinate actions. The settings are as follows:

\begin{itemize}
\item \textbf{Budget.} $B=300$ actions.
\item \textbf{Browser.} Playwright-controlled Chromium at $1280\times800$, a fresh
  browser context per episode.
\item \textbf{Simple harness.} A ReAct loop that observes the last three screenshots, persistent notes, visited pages, and available auxiliary actions, and outputs pixel coordinates.
\item \textbf{Claude Code.} Medium reasoning effort, with tools restricted to GUI actions and \texttt{report}.
\item \textbf{Decoding.} Temperature 0, except for Qwen 3.7 Plus, for which we use 0.3 because the API used in our experiments returns empty responses at temperature 0.
\item \textbf{Thinking.} Hidden reasoning is disabled wherever the provider
  exposes a switch (DeepSeek, GLM, Qwen, Doubao): it consumes each step's output
  budget and wall-clock time, and Qwen's reasoning mode frequently produced
  malformed action JSON. Gemini and Claude keep their provider defaults, and
  Claude Code runs at reasoning effort medium.
\item \textbf{Retries.} An empty reply is retried with a temperature bump up to 0.7;
  the request timeout is 120\,s.
\end{itemize}

\subsection{Extended Results}
\label{sec:supp_extended}

Table~\ref{tab:supp_extended} extends the main results with Claude Code runs
using DeepSeek-V4.1-Flash and GLM-5.3-Flash, and a simple-harness run using
DeepSeek-V4.1-Flash.

\tightpara{Harness benefits depend on the model}
Switching to Claude Code yields very different Recall changes across models:
+30.5 points for Opus, compared with $-3.2$ for DeepSeek-V4.1-Flash and +4.3
for GLM-5.3-Flash. Coverage and Reach improve for all three, suggesting that
Claude Code supports more effective exploration even when detection gains are
limited. Turning that exploration into useful reports appears to depend on
how well the model works with the harness. Possible reasons include closer integration between Claude Code and its native
model family and differences in the models' ability to use history and tool feedback.

\tightpara{Comparison within the same model family}
A stronger model does not necessarily achieve higher defect Recall. Under the simple harness, DeepSeek-V4-Pro and V4.1-Flash achieve the same overall Recall, with Flash performing better on display defects and Pro performing better on UTR and NLE defects. Possible explanations include differences in exploration and attention allocation, as well as Pro's stronger reasoning ability when checking complex interaction outcomes.

\begin{table*}[t]
  \centering
  \small
  \setlength{\tabcolsep}{4pt}
  \caption{\textbf{Extended results on GUITAR-Bench.} Same format as
  Table~1 of the main paper.}
  \label{tab:supp_extended}
  \begin{tabular}{l cccc ccccc cc cc}
    \toprule
    & \multicolumn{4}{c}{Overall} & \multicolumn{5}{c}{Category} & \multicolumn{2}{c}{Depth} & \multicolumn{2}{c}{Distance} \\
    \cmidrule(lr){2-5} \cmidrule(lr){6-10} \cmidrule(lr){11-12} \cmidrule(lr){13-14}
    Model & Cov. & Reach & Recall & Rec.$|$R. & EL & CR & ONR & UTR & NLE & Shallow & Deep & Near & Far \\
    \midrule
    \multicolumn{14}{l}{\textit{Claude Code}} \\
    \quad Opus 5.5 & 80.0 & 94.7 & 72.6 & 76.7 & 90.9 & 76.5 & 63.2 & 64.5 & 66.7 & 72.2 & 73.9 & 74.7 & 50.0 \\
    \quad DeepSeek-V4.1-Flash & 66.7 & 86.3 & 24.2 & 26.8 & 13.6 & 47.1 & 36.8 & 12.9 & 16.7 & 25.0 & 21.7 & 25.3 & 12.5 \\
    \quad GLM-5.3-Flash & 68.9 & 86.3 & 23.2 & 26.8 & 4.5 & 35.3 & 26.3 & 22.6 & 50.0 & 29.2 & 4.3 & 23.0 & 25.0 \\
    \midrule
    \multicolumn{14}{l}{\textit{Simple harness (ours)}} \\
    \quad Opus 5.5 & 68.6 & 83.2 & \textbf{42.1} & \textbf{49.4} & \textbf{45.5} & \textbf{52.9} & \textbf{57.9} & 25.8 & 33.3 & \textbf{40.3} & \textbf{47.8} & \textbf{43.7} & \textbf{25.0} \\
    \quad DeepSeek-V4-Pro & 65.1 & 82.1 & 27.4 & 33.3 & 22.7 & 23.5 & 26.3 & \textbf{29.0} & \textbf{50.0} & 27.8 & 26.1 & 28.7 & 12.5 \\
    \quad DeepSeek-V4.1-Flash & 59.8 & 78.9 & 27.4 & 34.7 & 31.8 & 41.2 & 31.6 & 16.1 & 16.7 & 29.2 & 21.7 & 28.7 & 12.5 \\
    \quad Gemini 3.8 Flash & \textbf{74.0} & \textbf{87.4} & 25.3 & 28.9 & 27.3 & 29.4 & 36.8 & 16.1 & 16.7 & 25.0 & 26.1 & 26.4 & 12.5 \\
    \quad GLM-5.3-Flash & 61.3 & 78.9 & 18.9 & 24.0 & 22.7 & 5.9 & 10.5 & 25.8 & 33.3 & 19.4 & 17.4 & 18.4 & \textbf{25.0} \\
    \quad Qwen3.7-Plus & 52.8 & 74.7 & 13.7 & 18.3 & 18.2 & 5.9 & 26.3 & 9.7 & 0.0 & 12.5 & 17.4 & 14.9 & 0.0 \\
    \quad Doubao-Seed-2.1-Turbo & 58.9 & 71.6 & 6.3 & 8.8 & 4.5 & 5.9 & 10.5 & 3.2 & 16.7 & 6.9 & 4.3 & 6.9 & 0.0 \\
    \bottomrule
  \end{tabular}
\end{table*}

\subsection{Recall Variability Across Apps and Runs}
\label{sec:supp_variability}

\tightpara{Recall variability under app bootstrap}
Table~\ref{tab:supp_recall_ci} reports 95\% percentile confidence intervals
from 10,000 bootstrap resamples on the apps. We sample
20 apps with replacement, retain their defects and detections, and
recompute pooled Recall. The CI summarizes how Recall varies when the mix of
apps changes, with recorded trajectories and judge decisions fixed. Claude Code with Opus remains
clearly separated from the other systems, while the uncertainty among
lower-scoring agents cautions against emphasizing small ranking differences.
The DeepSeek variants also illustrate that equal Recall can hide different
sensitivity to application composition: Flash has a wider interval than Pro.
These individual intervals do not constitute pairwise significance tests.

\begin{table}[t]
  \centering
  \small
  \caption{\textbf{Recall confidence intervals (CI) from app bootstrap.} We compute 95\% CIs from 10,000 bootstrap resamples of the 20 apps. Wider intervals indicate greater
  sensitivity of Recall to which apps are included.}
  \label{tab:supp_recall_ci}
  \setlength{\tabcolsep}{4pt}
  \begin{tabular}{lcc}
    \toprule
    Agent & Recall & 95\% CI \\
    \midrule
    \multicolumn{3}{l}{\textit{Claude Code}} \\
    Opus 5.5 & 72.6 & [64.2, 80.4] \\
    DeepSeek-V4.1-Flash & 24.2 & [15.2, 34.2] \\
    GLM-5.3-Flash & 23.2 & [14.6, 32.9] \\
    \midrule
    \multicolumn{3}{l}{\textit{Simple harness}} \\
    Opus 5.5 & 42.1 & [30.5, 52.9] \\
    DeepSeek-V4-Pro & 27.4 & [21.2, 33.3] \\
    DeepSeek-V4.1-Flash & 27.4 & [16.8, 39.2] \\
    Gemini 3.8 Flash & 25.3 & [17.5, 33.7] \\
    GLM-5.3-Flash & 18.9 & [11.2, 28.1] \\
    Qwen3.7-Plus & 13.7 & [7.0, 21.2] \\
    Doubao-Seed-2.1-Turbo & 6.3 & [2.4, 10.2] \\
    \bottomrule
  \end{tabular}
\end{table}

\tightpara{Recall variability across multiple runs}
We evaluate DeepSeek-V4.1-Flash and GLM-5.3-Flash four times each under the
simple harness(\Cref{tab:supp_recall_at_k}). Their Recall is
$21.1\pm4.9\%$ and $24.7\pm5.2\%$, respectively (\emph{mean} $\pm$ \emph{standard
deviation}). Although DeepSeek scores higher in the single-run comparison in
Table~\ref{tab:supp_extended}, GLM has a higher mean across the four runs.
This suggests some caution when interpreting small single-run differences
between models with similarly low recall. The single-run tables characterize
observed performance under a common protocol, rather than establish a definitive
ordering among closely performing agents. Our analysis emphasizes broader
detection patterns and the behaviors involved in reaching, triggering, and
recognizing defects.

\begin{table}[t]
  \centering
  \small
  \setlength{\tabcolsep}{5pt}
  \caption{\textbf{Per-run Recall and combined results across four runs.} Each agent is evaluated in four independent runs.
  Recall@$k$ averages union Recall over all $k$-run subsets of these four runs,
  counting each detected defect only once per subset}
  \label{tab:supp_recall_at_k}
  \begin{tabular}{lcc}
    \toprule
    & DeepSeek-V4.1-Flash & GLM-5.3-Flash \\
    \midrule
    Run 1 & 27.4 & 18.9 \\
    Run 2 & 22.1 & 24.2 \\
    Run 3 & 15.8 & 31.6 \\
    Run 4 & 18.9 & 24.2 \\
    Mean $\pm$ std & $21.1\pm4.9$ & $24.7\pm5.2$ \\
    \midrule
    Recall@1 & 21.1 & 24.7 \\
    Recall@2 & 29.3 & 34.9 \\
    Recall@3 & 33.7 & 40.8 \\
    Recall@4 & 35.8 & 45.3 \\
    \bottomrule
  \end{tabular}
\end{table}

\tightpara{Additional runs recover missed defects with diminishing gain} Using the same four runs, we examine how many defects missed in a single execution can be recovered through additional executions. We measure Recall@$k$ by taking the union of defects detected in each subset of $k$ runs, counting each defect once, and averaging the resulting recall over all such subsets of the four executions. This metric captures the additional discoveries enabled by variation across runs. As $k$ increases from one to four, Recall rises from 21.1\% to 35.8\% for DeepSeek and from 24.7\% to 45.3\% for GLM, but with diminishing gains: the second run adds 8.2 and 10.2 percentage points, respectively, whereas the fourth adds only 2.1 and 4.5 points. Nevertheless, Recall remains low for both models after four runs, with each still missing more than half of the defects, leaving substantial room for improvement.

\subsection{Precision and False Positive Analysis}
\label{sec:supp_precision}

\Cref{tab:supp_precision} complements Recall with Precision and $F_1$. Precision is the fraction of agent reports matched to injected
defects, and $F_1$ is the harmonic mean of precision and Recall. We first examine whether higher Recall comes from reporting more suspected
issues, then analyze the validity of unmatched reports in Section~\ref{sec:supp_unmatched}.

\begin{table}[t]
  \centering
  \small
  \setlength{\tabcolsep}{4pt}
  \caption{\textbf{Precision, Recall, and $F_1$.} Incorporating Precision through $F_1$ broadly preserves the ranking induced by Recall.
}
  \label{tab:supp_precision}
  \begin{tabular}{lrrr}
    \toprule
    Model & Precision & Recall & $F_1$ \\
    \midrule
    \multicolumn{4}{l}{\textit{Claude Code}} \\
    Opus 5.5 & 31.2 & 72.6 & 43.7 \\
    DeepSeek-V4.1-Flash & 12.5 & 24.2 & 16.5 \\
    GLM-5.3-Flash & 4.3 & 23.2 & 7.3 \\
    \midrule
    \multicolumn{4}{l}{\textit{Simple harness}} \\
    Opus 5.5 & 11.0 & 42.1 & 17.4 \\
    DeepSeek-V4-Pro & 7.9 & 27.4 & 12.3 \\
    DeepSeek-V4.1-Flash & 10.0 & 27.4 & 14.6 \\
    Gemini 3.8 Flash & 9.4 & 25.3 & 13.7 \\
    GLM-5.3-Flash & 5.6 & 18.9 & 8.7 \\
    Qwen3.7-Plus & 7.3 & 13.7 & 9.5 \\
    Doubao-Seed-2.1-Turbo & 2.4 & 6.3 & 3.5 \\
    \bottomrule
  \end{tabular}
\end{table}

\subsubsection{Does Higher Recall Come from More Reporting?}
\label{sec:supp_selectivity}

The main paper focuses on Recall; here we examine whether higher Recall is
simply achieved by reporting more suspected issues. As shown in ~\cref{tab:supp_precision}, Precision does
reveal differences that Recall alone misses. The two simple-harness DeepSeek
variants detect the same number of injected defects, but Flash has a higher
fraction of reports matching those defects (10.0 versus 7.9). Similarly,
simple-harness Qwen has lower Recall than GLM but higher precision, showing
that detecting more defects need not imply more selective reporting.

Despite these differences, Recall and $F_1$ give broadly consistent performance
rankings: Claude Code with Opus leads under both metrics, and Opus remains the
strongest model within the simple harness. Claude Code with Opus also submits
fewer reports than simple-harness Opus (237 versus 383), yet achieves both
higher Recall and higher precision. Its advantage therefore does not come
from more aggressive reporting. Accounting for precision thus preserves the
main paper's central findings, while refining comparisons among similarly
scoring models. This supports using Recall as the primary metric in the main paper: it directly
measures coverage of the verified target defects, and the leading results are
not an artifact of excessive reporting.

\subsubsection{Distinguishing Out-of-List Discoveries from False Positives}
\label{sec:supp_unmatched}

To distinguish out-of-list discoveries, i.e., valid defects present in the original app but not injected by us, from false positives, we audit 20 unmatched reports per agent by sampling 10 from each real/mock pool. We assess their validity through trajectory replay on the clean and defective applications, supported by source-code inspection.

\Cref{tab:supp_valid_discoveries} shows that agents can discover out-of-list defects, with Claude Code using Opus achieving the highest discovery rate. Although more out-of-list defects are found in mock applications, some are also found in real applications. Their presence in both pools illustrates the difficulty of establishing an exhaustive defect inventory in advance~\citep{barr2014oracle,hong2026framework}. Therefore, our benchmark instead provides a fixed set of known injected defects, validated through differential replay, as shared and reproducible evaluation targets. At the same time, Recall on injected defects remains generally low, indicating that GUI agents are still far from reliably finding the target defects. Thus, at this early stage, Recall should be prioritized, while Precision serves as a complementary measure of report selectivity and should receive greater emphasis once Recall becomes substantially higher.

\begin{table}[t]
  \centering
  \small
  \setlength{\tabcolsep}{4pt}
  \caption{\textbf{Out-of-list discovery rate.} To distinguish out-of-list discoveries, i.e., valid defects present in the original app but not injected by us, from false positives, we audit 20 unmatched reports per agent by sampling 10 from each real/mock pool. Validity is assessed through trajectory replay and source-code inspection.}
  \label{tab:supp_valid_discoveries}
  \begin{tabular}{lrrr}
    \toprule
    Model & Real & Mock & Rate (\%) \\
    \midrule
    \multicolumn{4}{l}{\textit{Claude Code}} \\
    Opus 5.5 & 3/10 & 8/10 & 55 \\
    DeepSeek-V4.1-Flash & 5/10 & 3/10 & 40 \\
    \midrule
    \multicolumn{4}{l}{\textit{Simple harness}} \\
    DeepSeek-V4.1-Flash & 1/10 & 0/10 & 5 \\
    GLM-5.3-Flash & 1/10 & 3/10 & 20 \\
    Qwen3.7-Plus & 1/10 & 1/10 & 10 \\
    \bottomrule
  \end{tabular}
\end{table}
\clearpage
\end{document}